\documentclass[lettersize,journal]{IEEEtran}
\usepackage{amsmath,amsfonts}
\usepackage{algorithmic}
\usepackage{algorithm}
\usepackage{array}
\usepackage[caption=false,font=normalsize,labelfont=sf,textfont=sf]{subfig}
\usepackage{textcomp}
\usepackage{stfloats}
\usepackage{url}
\usepackage{verbatim}
\usepackage{graphicx}
\usepackage{cite}
\usepackage{ragged2e} 
\usepackage{booktabs,makecell, multirow, tabularx}
\usepackage{bigstrut}
\usepackage{graphicx}
\usepackage{bbding}
\usepackage{blindtext}
\usepackage{setspace}
\usepackage{algorithmic}
\usepackage{algorithm}

\usepackage[table,xcdraw]{xcolor}

\begin{document}

\title{GraphMamba: An Efficient Graph Structure Learning Vision Mamba for Hyperspectral Image Classification}

\author{Aitao Yang, Min Li, Yao Ding, Leyuan Fang,~\IEEEmembership{Senior Member,~IEEE,} Yaoming Cai,~\IEEEmembership{Member,~IEEE,} and Yujie He

\thanks{This work was supported in part by the National Natural Science Foundation of China under Grant 62006240. (Corresponding authors: Min Li; Yao Ding.)}

\thanks{Aitao Yang, Min Li, Yao Ding, Yujie He are with the Xi’an Research Institute of High Technology, Xi’an 710025, China. (e-mail: 824360083@qq.com; proflimin@163.com; dingyao.88@outlook.com; ksy5201314@163.com.).}

\thanks{Leyuan Fang is with the College of Electrical and Information Engineering, Hunan University, Changsha 410082, China, and also with the Peng Cheng Laboratory, Shenzhen 518000, China (e-mail: fangleyuan@gmail.com).}

\thanks{Yaoming Cai is with the School of Information Engineering, Zhongnan University of Economics and Law, Wuhan 430073, China, and also with the Emergency Management Research Center, Zhongnan University of Economics and Law, Wuhan 430073, China (e-mail: caiyaom@zuel.edu.cn).}

}

\maketitle

\begin{abstract}
Efficient extraction of spectral sequences and geospatial information has always been a hot topic in hyperspectral image classification. In terms of spectral sequence feature capture, RNN and Transformer have become mainstream classification frameworks due to their long-range feature capture capabilities. In terms of spatial information aggregation, CNN enhances the receptive field to retain integrated spatial information as much as possible. However, the spectral feature capturing architectures exhibit low computational efficiency, and CNNs lack the flexibility to perceive spatial contextual information. To address these issues, this paper proposes GraphMamba—an efficient graph structure learning vision Mamba classification framework that fully considers HSI characteristics to achieve deep spatial-spectral information mining. Specifically, we propose a novel hyperspectral visual GraphMamba processing paradigm (HVGM) that preserves spatial-spectral features by constructing spatial-spectral cubes, and utilize linear spectral encoding to enhance the operability of subsequent tasks. The core components of GraphMamba include the HyperMamba module for improving computational efficiency and the SpectralGCN module for adaptive spatial context awareness. The HyperMamba mitigates clutter interference by employing the global mask (GM) and introduces a parallel training inference architecture to alleviate computational bottlenecks. The SpatialGCN incorporates weighted multi-hop aggregation (WMA) spatial encoding to focus on highly correlated spatial structural features, thus flexibly aggregating contextual information while mitigating spatial noise interference. It is worth noting that the encoding modules of the proposed GraphMamba architecture are flexible and scalable, providing a new approach for joint mining of HSI spatial-spectral information. Extensive experiments were conducted on three different scales of real HSI datasets, and compared with the state-of-the-art classification frameworks, GraphMamba achieved optimal performance. The core
code will be released at https://github.com/ahappyyang/GraphMamba.

\end{abstract}


\section{Introduction}

\IEEEPARstart{H}{yperspectral} images (HSIs) consist of many narrow and contiguous spectral bands. Compared to traditional panchromatic and multispectral remote sensing images, they have the advantage of a wide spectral response range and high spectral resolution. Additionally, the hyperspectral image exhibits strong capabilities in target coverage identification, as hyperspectral data can detect materials with diagnostic spectral absorption features, accurately distinguishing land cover types, road pavement materials, and more. Countries worldwide attach great importance to the development of hyperspectral remote sensing, with its applications becoming increasingly widespread, achieving remarkable results in the environmental monitoring~\cite{environmental-monitoring}, land cover classification~\cite{land-cover}, military target recognition~\cite{military}, and precision agriculture~\cite{precision-agriculture} field.

Hyperspectral images contain rich information about land covers. However, the abundance of spectral bands leads to information redundancy and high inter-band correlation, resulting in the Hughes phenomenon. Furthermore, hyperspectral images have low spatial resolution, leading to mixed pixel effects. Relying solely on spectral information for representation and learning can potentially cause misclassification. These challenges make the efficient processing of massive information and accurate classification of objects more demanding.

To address the deficiencies of spectral information representation in supervised classification, researchers have proposed many methods for HSI classification in the past decade. These methods can be broadly categorized into two types: traditional methods and deep learning methods. Traditional methods focus on discriminative feature extraction and compact representation of spectral data, with representative methods including random forests~\cite{RF}, K-nearest neighbors (KNN)~\cite{K-means}, and support vector machines (SVM)~\cite{ISPRS1}. However, these methods always classify pixels in HSIs as independent spectral curves, neglecting spatial structural information, which leads to low classification accuracy. Additionally, it lacks robustness in the presence of noise and information loss.

\begin{figure*}
\centerline{\includegraphics[width=0.93\textwidth]{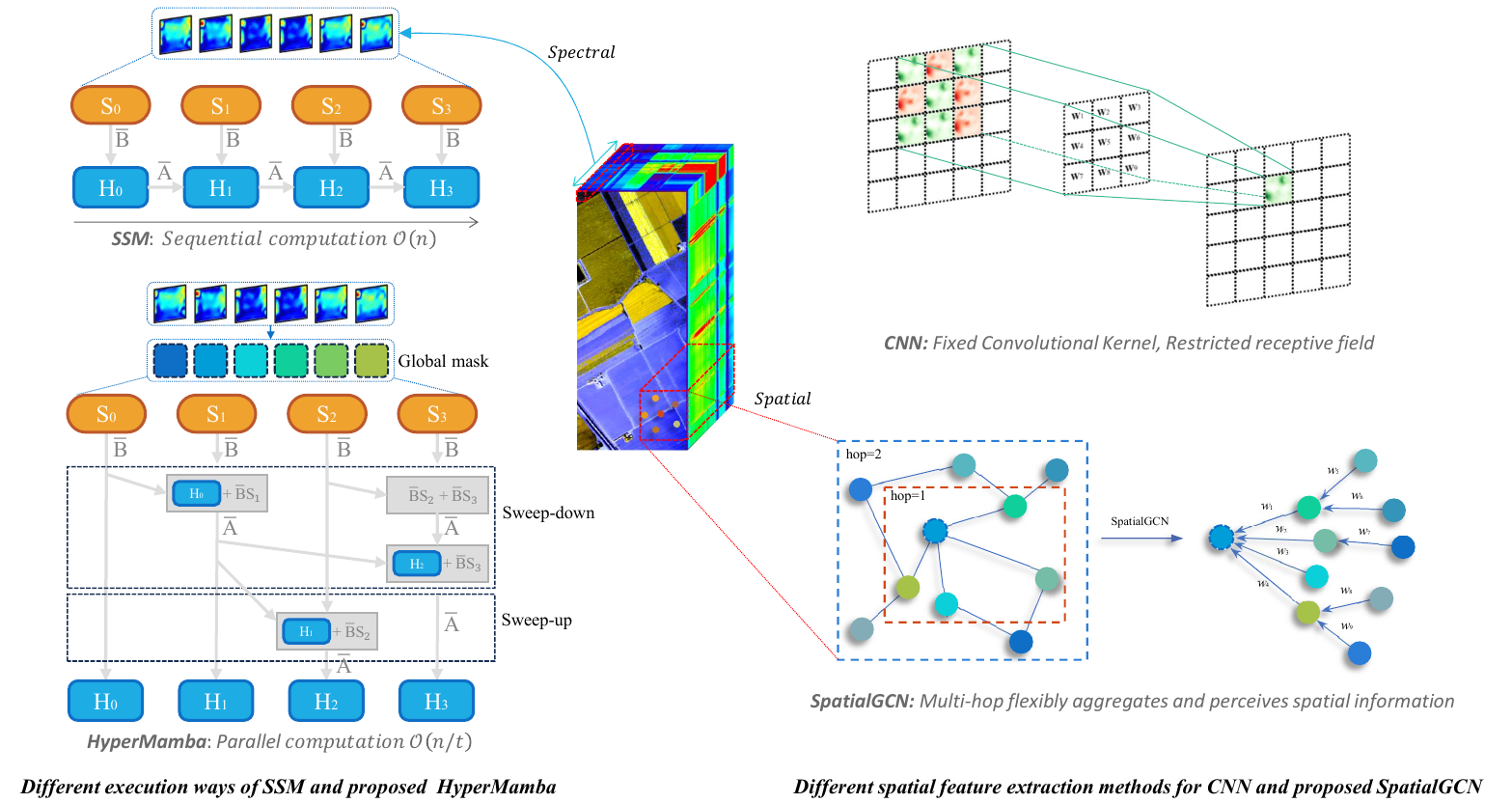}} 
\caption{The improvements of the proposed spatial-spectral feature extraction modules HyperMamba and SpatialGCN compared to other frameworks are discussed in this paper.}
\label{fig:0}
\end{figure*}

In recent years, with the rapid development of deep learning, its powerful nonlinear expressive capability has attracted wide attention. Compared to traditional methods, deep learning methods have achieved automatic extraction of high-level semantic information, thus avoiding cumbersome feature engineering. RNN is a type of network suitable for processing sequential data, which can effectively capture discriminative features in sequences. Considering that spectra are a type of sequence, RNN can fully utilize the high spectral resolution of HSI to capture long-range dependencies of features. For example, Zhou \MakeLowercase{\textit{et al.}}~\cite{RNN1} used cascaded RNNs to eliminate redundant information between spectral bands and simultaneously explored complementary information in spectral sequences. Zhang \MakeLowercase{\textit{et al.}}~\cite{RNN2} employed the local spatial sequential (LSS) method to extract low-level structural information, which was then passed to RNN for generating high-level semantic information. Compared to RNN, Transformer~\cite{ISPRS-TR} can better capture global dependencies in sequences and possess high flexibility. Mei \MakeLowercase{\textit{et al.}}~\cite{transformer} introduced the group embedding module in Transformer to achieve local-global spectral context feature extraction. To address the insufficient spatial feature exploration capability of Transformer, Yang \MakeLowercase{\textit{et al.}} proposed a joint GCN and Transformer network to capture subtle spectral differences. Moreover, Convolutional Neural Networks (CNNs)~\cite{ISPRS-CNN,ISPRS-CNN2} have been widely applied in HSI classification due to their excellent spatial perception ability. To tackle the challenges of spectral information redundancy and insufficient spatial resolution, Chang \MakeLowercase{\textit{et al.}}~\cite{CNN1} designed a comprehensive CNN, where 2-D CNN was used to extract spatial semantic information, and 3-D CNN was responsible for extracting spectral discriminative features to reduce information redundancy. Ye \MakeLowercase{\textit{et al.}}~\cite{CNN2} creatively utilized the particle swarm algorithm to optimize the parameters of CNN, thereby enhancing the model's generalization performance and fast learning ability. Additionally, Autoencoders (AE), Generative Adversarial Networks (GANs), and Capsule Networks (CapsNets) have also been widely applied in HSI classification. However, the aforementioned networks face the following urgent issues that need to be addressed:

\begin{enumerate}
\item{RNNs have high computational complexity during training and inference, limiting their scalability in handling long sequences and large-scale data.}

\item{As the sequence length increases, the computational complexity of Transformers grows quadratically, becoming a bottleneck that restricts its efficiency.}

\item{The weights of CNN kernels are fixed and cannot adaptively capture spatial structural changes.}

\end{enumerate}

\begin{figure*}
\centerline{\includegraphics[width=0.97\textwidth]{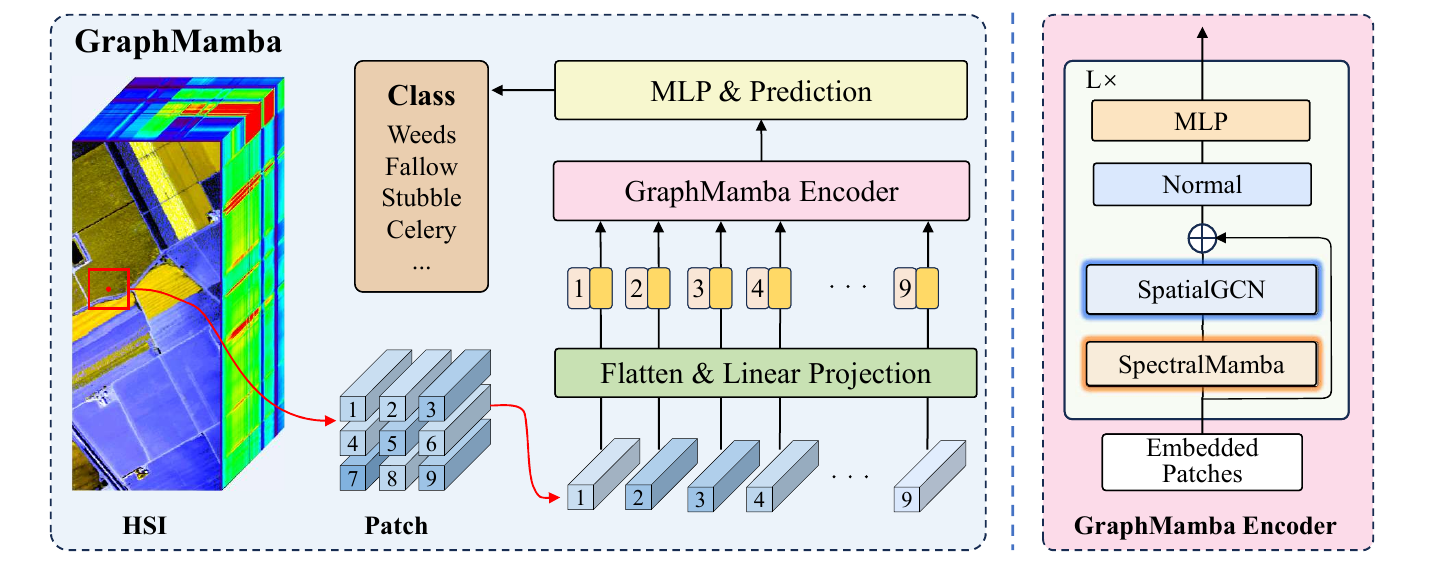}} 
\caption{The overall view of GraphMamba. We first segment HSI into multiple spatial-spectral cubes, then linearly project them into patch tokens, and then send the token sequences to the proposed GraphMamba encoders to extract features. Finally, classification prediction features are obtained through MLP.}
\label{fig:1}
\end{figure*}

In recent years, benefiting from the development of State Space Models (SSM)~\cite{SSM}, which alleviates the problem of low computational efficiency in RNNs and Transformers, a new perspective for modeling sequence information has been provided. Additionally, the computational complexity and memory usage of SSM are linearly related to the sequence length, which is crucial for enabling fast training and inference. Furthermore, to address the long-range dependency modeling issue in limited storage space with SSM, Gu \MakeLowercase{\textit{et al.}}~\cite{HiPPO} introduced HiPPO, which compresses input signals into coefficient vectors and constructs a new memory update mechanism to retain all history. For performing inference calculations on ultra-long sequences, the Structured State Space sequence model (S4)~\cite{S4} was proposed, further simplifying SSM to the computation of a Cauchy kernel. Recently, a new SSM-based framework called Mamba~\cite{mamba} has been introduced. In language and audio tasks~\cite{ViM}, Mamba has been shown to surpass Transformers of equivalent scale. Its selective information processing, simpler design architecture, and parallel scanning algorithm make it more efficient, with broad application prospects. The rapid development of SSM has revealed its enormous potential and high adaptability in sequence data mining. However, the high-dimensional spectral data and information redundancy pose challenges for the application of SSM in HSI classification.

In capturing dynamic spatial structures, Graph Convolutional Networks (GCN) are more flexible compared to CNN. GCN can extract spatial features of topological graphs of non-Euclidean geometry structures and explicitly aggregate node features based on the similarity between pixels to construct adjacency matrices. To help GCNs better understand global contextual new information, Ding \MakeLowercase{\textit{et al.}}~\cite{GCN1} employed a multiscale graph aggregation network, graphSAGE, to automatically learn deep semantic information of hyperspectral imagery (HSI). Xu \MakeLowercase{\textit{et al.}}~\cite{GCN2} optimized GCN using fuzzy theory to build robust graphs for better representing complex spatial-spectral relationships in HSI.

In response to the aforementioned challenges and achieve efficient parallel computation while adaptively mining spatial feature relationships, we thoroughly explore the potential of SSM in HSI classification and propose an efficient graph structure learning vision mamba for hyperspectral image classification—GraphMamba. GraphMamba incorporates a series of optimization designs tailored to the characteristics of HSI, serving as an end-to-end efficient spatial-spectral joint feature extraction classification network. Specifically, to reduce training resource consumption for efficient computation, we introduce a novel image processing paradigm, hyper-vision mamba, specifically designed for HSI. Additionally, the global mask is introduced to refine high-dimensional data and HyperMamba is proposed to mine deep semantic information. Finally, we utilize multi-hop neighborhood aggregation matrices to better capture global graph structural information and further construct SpatialGCN for spatial graph structure learning. The main contributions of this paper are as follows:

\begin{enumerate}
\item{Introducing a novel graph structure learning vision Mamba paradigm, achieving efficient representation of deep spatial-spectral features through the adaptive GraphMamba encoder.}

\item{Proposing HyperMamba, which reduces information redundancy interference by introducing global spectral masks and decreases computational costs through parallel scanning.}

\item{Designing SpatialGCN with enhanced context-aware capabilities. The proposal of multi-hop adaptive neighborhood matrices enables better extraction of dynamic spatial structural features.}

\end{enumerate}

\section{Review of SSM and GCN}

\subsection{State Space Models}

SSM refers to State Space Models(SSM), a type of sequence model used in deep learning. SSM maps the implicit latent state $h(t) \in {\mathbb{R}^N}$ to the feature sequence $x(t) \in \mathbb{R} \to y(t) \in \mathbb{R}$. Deep networks abstracted from SSM can overcome the issue of Transformer being insensitive to discrete deep features in sequence prediction tasks, thereby establishing stronger logic for temporal data.

The key components of SSM are the state representation equation and the prediction output equation. The continuous SSM representation is as follows:
\begin{equation}
{h^\prime }(t) = {\bf{A}}h(t) + {\bf{B}}x(t)
\end{equation}
\begin{equation}y(t) = {\bf{C}}h(t)\end{equation}
where $\bf{A}$ represents the state transition matrix, and $\bf{B}$ and $\bf{C}$ respectively represent the mapping matrices from input to latent state and from latent state to output, which remains fixed and unchanged.

In image processing tasks, signals are often discrete rather than continuous. To better handle discrete sequences, continuous SSM needs to be transformed into discrete SSM:
\begin{equation}
{h_k} = \overline {\bf{A}} {h_{k - 1}} + \overline {\bf{B}} {x_k}
\end{equation}
\begin{equation}{y_k} = \overline {\bf{C}} {h_k}\end{equation}
\begin{equation}\overline {\bf{A}}  = {e^{\Delta {\rm{A}}}}\end{equation}
\begin{equation}\overline {\bf{B}}  = {(\Delta {\bf{A}})^{ - 1}} \cdot ({e^{\Delta {\bf{A}}}} - {\bf{I}}) \cdot \Delta {\bf{B}}\end{equation}
\begin{equation}\overline {\bf{C}}  = {\bf{C}}\end{equation}
where $k$ represents the step size, $\overline {\bf{A}} $, $\overline {\bf{B}} $, and $\overline {\bf{C}} $ represent the discrete matrices obtained through the Zero-order hold technique. Further obtain the mapping representation ${y_k}$ of the discrete sequence ${x_k}$.

\subsection{Graph Convolutional Network}
Graph convolutional networks(GCN) can represent HSI as undirected graphs,  thereby establishing relationships between land covers. For a graph ${\cal G} = ({\cal V},{\cal E},{\bf{A}})$, ${\cal V}$ represents a set of vertices, ${\cal E}$ is the edge set, ${\bf{A}}\in {{\mathop{\rm R}\nolimits} ^{{\mathop{\rm N}\nolimits} \times N}}$ is the adjacency matrix of ${\cal G}$. Given the adjacency matrix $\bf{A}$, the Laplace matrix $\bf{L}$ of the ${\cal G}$ is:
\begin{equation}
{\bf{L}} = {\bf{D}} - {\bf{A}}
\end{equation}
where $\bf{D}$ is the degree matrix of ${\cal G}$.

Normalizing the Laplace matrix $\bf{L}$ yields:

\begin{equation}
{{\bf{L}}_n} = {\bf{I}} - {{\bf{D}}^{\frac{1}{2}}}{\bf{A}}{{\bf{D}}^{ - \frac{1}{2}}}
\end{equation}
where $\bf{I}$ is the identity matrix.

GCN introduces a convolutional kernel ${g_\theta } = diag(\theta )$ to embed nodes, where $\theta \in {R^N}$. The formula for graph convolution can be expressed as:

\begin{equation}
{x_G} * {g_\theta } = {\bf{U}}{g_\theta }(\Lambda ){{\bf{U}}^T}x
\end{equation}
where ${x_G}$ is the node representation, $\bf{U}$ is the eigenvector matrix of ${{\bf{L}}_n}$, $\Lambda $ is the eigenvalues of ${{\bf{L}}_n}$. To simplify the calculation, the Chebyshev polynomial is used to fit ${g_\theta }(\Lambda )$:

\begin{equation}
{g_\theta }(\Lambda ) \approx \sum\limits_{k = 0}^K {{\theta _K}{T_K}({{\bf{L}}_n})}
\end{equation}
where ${T_K}$ is the Chebyshev polynomial.

Taking the first-order Chebyshev polynomial for further simplification, we can obtain:

\begin{equation}
{x_G} * {g_\theta } = \theta ({{\bf{\tilde D}}^{\frac{1}{2}}}{\bf{\tilde A}}{{\bf{\tilde D}}^{ - \frac{1}{2}}})x
\label{eq:10}
\end{equation}


Based on Eq.\!~\ref{eq:10}, the propagation rule of GCN is as follows:

\begin{equation}
{{\bf{H}}^{(l + 1)}} = \sigma ({{\bf{\tilde D}}^{\frac{1}{2}}}{\bf{\tilde A}}{{\bf{\tilde D}}^{ - \frac{1}{2}}}{{\bf{H}}^{(l)}}{{\bf{W}}^{(l)}})
\end{equation}
where ${{\bf{H}}^{(l + 1)}}$ and ${{\bf{H}}^{(l)}}$ represent the features of the $l+1$ and $l$ layers respectively, $\sigma(\cdot)$ represents the activation function, ${\bf{W}}$ represents the weight matrix.

\section{Proposed Methodology}

In this section, we first introduce a novel HSI processing paradigm HVGM, which establishes a spatial-spectral cube to retain local information while alleviating the input volume through block-wise input. Next, to address the issues of spectral redundancy and high computational resource consumption, we design the HyperMamba module to introduce GM for selective extraction of spectral features and parallel computation. Additionally, we introduce the SpectialGCN module to achieve adaptive spatial information mining by aggregating neighborhood information through WMA. Finally, an analysis of model complexity is conducted.

\subsection{GraphMamba}

The HSI image distinguishes land cover types by differences in pixel values (reflecting the spectral information) and spatial variations (indicating the spatial information). Compared to RNN and Transformer, Mamba utilizes SSM to dynamically filter and process information based on inputs, allowing the model to selectively remember or ignore input parts, thus enabling more effective processing of spectral sequences.

This paper aims to develop a universal and scalable HSI classification baseline network based on Mamba. To achieve this, we propose a novel HSI processing paradigm HVGM by partitioning HSI into spatial-spectral cubes to retain spectral information while reducing computational burdens. The overall architecture of GraphMamba is illustrated in Fig.\!~\ref{fig:1}.

Mamba usually encodes one-dimensional sequences as tokens. To better adapt to hyperspectral image classification, we first unfold the three-dimensional image ${\bf{h}} \in {\mathbb{R}^{H \times W \times D}}$ into $N$ two-dimensional patches ${{\bf{x}}} \in {\mathbb{R}^{N \times ({P^2} \cdot D)}}$, where $H$, $W$ and $D$ represent the length, width, and spectral dimension of the hyperspectral image, and $P$ is the patch size. Then, we linearly project ${{\bf{x}}}$ into $C$ dimensions and introduce position vectors ${{\mathbf{F}}_{pos}} \in {\mathbb{R}^{N \times D}}$:

\begin{equation}{{\bf{M}}_0} = [{\bf{Wx}}_p^1;{\bf{Wx}}_p^1;{\bf{Wx}}_p^1; \cdot  \cdot  \cdot {\bf{Wx}}_p^N] + {{\bf{F}}_{pos}}
\label{eq:12}
\end{equation}
where ${\bf{x}}_p^n$ denotes the $n$-th patch of ${\bf{x}}$, and ${\bf{W}}$ is a learnable parameter matrix.
Then we input the features ${\bf{M}}_l$ into the encoding layer ${\rm{GrM}}(\cdot)$ of the l-th GraphMamba to obtain the output ${\bf{M}}_{l+1}$.

\begin{equation}{{\bf{M}}_{l{\rm{ + 1}}}}{\rm{ = GrM(}}{{\bf{M}}_l}{\rm{) + }}{{\bf{M}}_l}\end{equation}
\begin{equation}{{\bf{M'}}_c}{\rm{ = Norm(}}{\bf{M}}_{l + 1}^c{\rm{)}}\end{equation}
\begin{equation}{\bf{y}} = {\mathop{\rm MLP}\nolimits} ({{\bf{M'}}_c})\end{equation}
where ${\bf{M}}_{l + 1}^c$ represents the output features corresponding to the target land cover element $c$. We employ normalization ${\rm{Norm(}} \cdot {\rm{)}}$ and a multi-layer perceptron ${\mathop{\rm MLP}\nolimits} ( \cdot )$ to obtain the final classification vector $y$. It is worth noting that we do not use classification tokens but directly classify using element features to further simplify the network structure.

\subsection{HyperMamba}

\begin{figure}
\centerline{\includegraphics[width=0.43\textwidth]{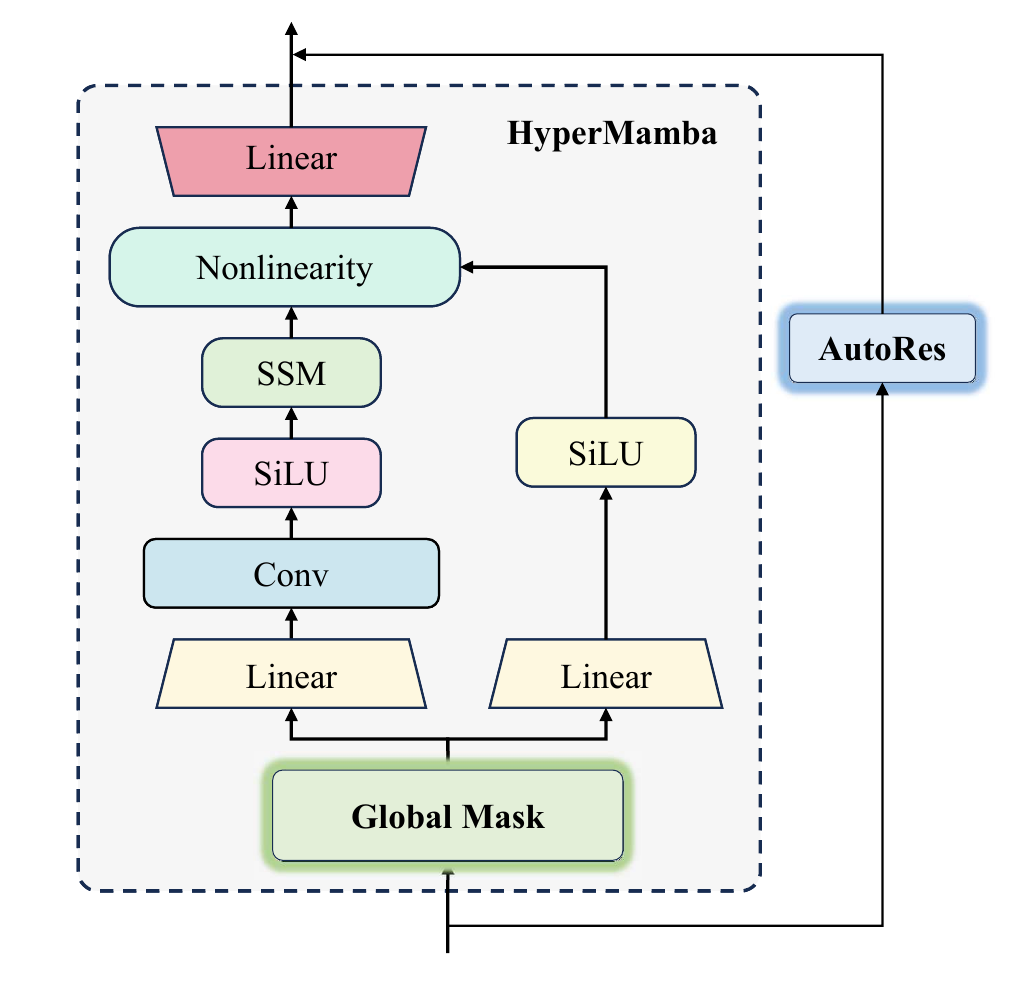}} 
\caption{Diagram of the HyperMamba in the proposed GraphMamba. The luminous marking modules are Global Mask and AutoRes, respectively.}
\label{fig:2}
\end{figure}

HSIs have high spectral dimensions and large data volume~\cite{ISPRS-HSI}. To efficiently establish a bridge for information exchange between different bands, focusing on more discriminative information, this paper proposes the Global Mask (GM) to aggregate spectral information based on the correlation between bands, while alleviating the issue of information redundancy. Assuming the input features are denoted by ${\mathbf{S}} \in {\mathbb{R}^{1 \times D}}$, the computation process of the Global Mask is as follows:

\begin{equation}{\bf{\hat S}} = Conv({{\bf{R}}_s} \cdot {\bf{C}}_s^T){\bf{S}}
\label{eq:16}
\end{equation}
\begin{equation}{{\bf{R}}_s} = {{\bf{W}}_R} \cdot {\bf{S}}
\label{eq:17}
\end{equation}
\begin{equation}{{\bf{C}}_s} = {{\bf{W}}_C} \cdot {\bf{S}}
\label{eq:18}
\end{equation}
where ${{\mathbf{W}}_R},{{\mathbf{W}}_C} \in {\mathbb{R}^{1 \times D}}$ represents learnable weight parameters, $Conv(\cdot)$ denotes 2-D convolution, $Conv(\textbf{R}_s \cdot \textbf{C}_s^T)$ forms a global mask, and consequently obtains the output ${\bf{\hat S}}$, which better filters relevant information and achieves efficient data utilization.

Residual connections are often used between different encoding blocks to promote information exchange and gradient propagation between networks, which can enhance the model's expressive power and feature utilization~\cite{RES}. However, directly using residual connections for HSI classification can easily introduce original spectral noise\cite{SF}. To reduce the model's sensitivity to noise, this paper designs an Automatic Residual Connection (AutoRes) to selectively incorporate original information. Assuming the input features of the $l$-th encoding layer are denoted by $\textbf{M}_l$, AutoRes is represented as follows:
\begin{equation}
\begin{array}{l}
{\bf{Z}} = {\rm{AutoRes}}({{\bf{M}}_l},{{\bf{M}}_{l - 1}})\\
 = \varepsilon {{\bf{M}}_l} + (1 - \varepsilon ){{\bf{M}}_{l - 1}}
\end{array}
\end{equation}
where $\varepsilon$ is a learnable parameter, and ${\bf{Z}}$ represents the output features. The automatic residual module is concise and efficient, as it adaptively integrates output features and original information, alleviating the issue of over-smoothing features.

\subsection{SpatialGCN}

GCN possesses strong spatial information aggregation capabilities and flexibility. However, the graphs are typically manually constructed, and the graphs used often struggle to represent global contextual information~\cite{GCN4}. Existing solutions often involve constructing multiple graph branches with different neighborhood ranges to expand the perceptual scope~\cite{GCN_db1}. However, this approach not only leads to a quadratic increase in computational complexity but also tends to introduce aberrant information.

To adaptively capture spatial contextual information, this paper proposes Weighted Multi-hop Aggregation Spatial Graph Convolutional Encoding (SpatialGCN) to further enhance the model's generalization ability. Assuming the input is ${S} \in {\mathbb{R}^{1 \times D}}$ , SpatialGCN can be represented as:

\begin{equation}{H^p} = \varphi (W\sum\limits_{n = 0}^p {({Q^n} \cdot {Z^n}))} 
\label{eq:20}
\end{equation}
\begin{equation}{Z^0} = S
\label{eq:21}
\end{equation}
\begin{equation}{Z^p} = {\bf{\tilde D}}_p^{\frac{1}{2}}{{\bf{\tilde A}}_p}{\bf{\tilde D}}_p^{ - \frac{1}{2}}S
\label{eq:22}
\end{equation}
where "$ \cdot $" denotes dot product, $Q$ represents the adaptive filtering matrix, $p$ represents the $p$-hop neighboring nodes of the node, ${Z^p} \in {\mathbb{R}^{1 \times D}}$ denotes the $p$-hop aggregated features of the node, and when $p$=0, ${Z^0} = S$. ${{\mathbf{A}}_p},{\mathbf{D}}_p^{} \in {\mathbb{R}^{D \times D}}$ respectively represent the adjacency matrix and the degree matrix of $n$-hop connections. $\varphi ( \cdot )$ denotes the activation function, and $W$ is the learnable shared parameter matrix. When $p$=1, ${H^p}$ is a normal GCN with residual connections, providing good scalability and adaptability.

\begin{algorithm}[!t]
	\renewcommand{\algorithmicrequire}{\textbf{Input:}}
	\renewcommand{\algorithmicensure}{\textbf{Output:}}
	\caption{Mechanism of GraphMamba}
	\label{algorithm: masp}
	\begin{algorithmic}[1]
		\REQUIRE  Hyperspectral image, learning rate $\eta=0.0001$, number of Epochs ${\cal T}$, patch size $p$;

        \STATE Divide the original HSI image into ${\cal N}$ patches of size $p \times p$;\\
// Train HyperMamba model
	   \FOR{$\tau = 1$ to ${\cal T}$}{

			\FOR{$n = 1$ to ${\cal N}$} {		

			        \STATE Introduce positional encoding based on formula\!~\ref{eq:12} and linearly process the input to obtain ${{\bf{M}}_0}$;\\
				    // HyperMamba filters information and extracts features.
			        \STATE Use Global Mask based on formulas\!~\ref{eq:16}-\!\!\!~\ref{eq:18} to filter spectral information in ${{\bf{M}}_0}$;
			        \STATE Input the filtered information into the HyperMamba module and extract spatial information through automatic residual connections;
				   \STATE Batch normalization, dropout, and ReLU;\\
					// SpatialGCN extracts spatial contextual information.
				   \STATE Construct an adaptive filtering matrix ${Q^n}$ based on formula\!~\ref{eq:23};
				   \STATE Perform context-aware learning based on formula\!~\ref{eq:20}-\!\!\!~\ref{eq:22};
				   \STATE Obtain the final output through normalization and MLP;
				   \STATE Calculate the loss using cross-entropy loss function, and update weight matrices using Adam gradient descent;
			}\ENDFOR

	  }\ENDFOR
	   \STATE Make predictions on test samples based on the trained network;
		
		\ENSURE
			Predict all pixel land cover classes $L$.
	\end{algorithmic}
\end{algorithm}

Additionally, the calculation of the adaptive filtering matrix ${Q^n}$ is as follows:

\begin{equation}{q_{ij}} = \left\{ {\begin{array}{*{20}{c}}
{\frac{{\exp ( - \gamma {{\left\| {{s_i} - {s_j}} \right\|}^2})}}{{\sum\limits_{j = 0}^n {\exp ( - \gamma {{\left\| {{s_i} - {s_j}} \right\|}^2})} }}{\kern 1pt} ,if{\kern 1pt} a_{ij}^p \ne 0}\\
{0,{\kern 1pt} {\kern 1pt} {\kern 1pt} if{\kern 1pt} a_{ij}^p = 0}
\end{array}} \right.
\label{eq:23}
\end{equation}
where ${q_{ij}}$ represents the element in the matrix ${Q^n}$, ${s_i}$ denotes the $i$-th node in the graph, $\gamma = 0.2$ is a constant, and ${a_{ij}^p}$ is the value at position $(i,j)$ in the adjacency matrix ${\bf{A}_p}$. Different from the simple way of adding branches in a multi-hop GCN, SpatialGCN adopts weight sharing to effectively reduce the parameter volume, thereby alleviating computational pressure. In addition, through adaptive filtering matrices, it can abstract the relevance of different nodes and better extract spatial structural features.

\subsection{ Computational complexity analysis}

Given a sequence of n spectral features $h \in {\mathbb{R}^{p \times d}}$, where $p$ represents the product of the height and width of a patch and $d$ is the dimension of the features after linear processing. The computational complexity of Global Mask is ${\cal O}(2n{p^2}d)$, the computational complexity of the SSM module is ${\cal O}(3p(2d)n + p(2d)n)$, and the computational complexity of GCN is ${\cal O}(|{\cal E}|{p^2}d)$, where $|{\cal E}|$ denotes the number of non-zero entries in the adjacency matrix $\bf{A}$. It can be observed that the computational complexity of SSM grows linearly with the sequence length and feature dimension, greatly enhancing computational efficiency.

\begin{table*}
\renewcommand\arraystretch{1}
\caption{The details of each land-cover class of three datasets used for evaluation in the experiments.}
\centering
\resizebox{\textwidth}{!}{%
\begin{tabular}{@{}ccccccccccccc@{}}
\toprule[1.5pt]
\multirow{2}{*}{NO.}       & \multicolumn{4}{c}{Indian Pines}                                               & \multicolumn{4}{c}{Salinas}                                              & \multicolumn{4}{c}{Houston 2013}             \\ \cline{2-13} 
                           & Class name                    & Train & Validation & Test                      & Class name             & Train & Validation & Test                       & Class name      & Train & Validation & Test  \\ \midrule
\multicolumn{1}{c|}{1}     & Alfalfa                       & 15    & 15         & \multicolumn{1}{c|}{31}   & Weeds\_1               & 30    & 30         & \multicolumn{1}{c|}{1979}  & Healthy grass   & 30    & 30         & 1221  \\
\multicolumn{1}{c|}{2}     & Corn-notill                   & 30    & 30         & \multicolumn{1}{c|}{1398} & Weeds\_2               & 30    & 30         & \multicolumn{1}{c|}{3696}  & Stressed grass  & 30    & 30         & 1224  \\
\multicolumn{1}{c|}{3}     & Corn-mintill                  & 30    & 30         & \multicolumn{1}{c|}{800}  & Fallow                 & 30    & 30         & \multicolumn{1}{c|}{1946}  & Synthetic grass & 30    & 30         & 667   \\
\multicolumn{1}{c|}{4}     & Corn                          & 30    & 30         & \multicolumn{1}{c|}{207}  & Fallow\_rough\_plow    & 30    & 30         & \multicolumn{1}{c|}{1364}  & Trees           & 30    & 30         & 1214  \\
\multicolumn{1}{c|}{5}     & Grass-pasture                 & 30    & 30         & \multicolumn{1}{c|}{453}  & Fallow\_smooth         & 30    & 30         & \multicolumn{1}{c|}{2648}  & Soil            & 30    & 30         & 1212  \\
\multicolumn{1}{c|}{6}     & Grass-trees                   & 30    & 30         & \multicolumn{1}{c|}{700}  & Stubble                & 30    & 30         & \multicolumn{1}{c|}{3929}  & Water           & 30    & 30         & 295   \\
\multicolumn{1}{c|}{7}     & Grass-pasture-mowed           & 15    & 15         & \multicolumn{1}{c|}{13}   & Celery                 & 30    & 30         & \multicolumn{1}{c|}{3549}  & Residential     & 30    & 30         & 1238  \\
\multicolumn{1}{c|}{8}     & Hay-windrowed                 & 30    & 30         & \multicolumn{1}{c|}{448}  & Grapes\_untrained      & 30    & 30         & \multicolumn{1}{c|}{11241} & Commercial      & 30    & 30         & 1214  \\
\multicolumn{1}{c|}{9}     & Oats                          & 15    & 15         & \multicolumn{1}{c|}{5}    & Soil\_vinyard\_develop & 30    & 30         & \multicolumn{1}{c|}{6173}  & Road            & 30    & 30         & 1222  \\
\multicolumn{1}{c|}{10}    & Soybeans-notill               & 30    & 30         & \multicolumn{1}{c|}{942}  & Corn                   & 30    & 30         & \multicolumn{1}{c|}{3248}  & Highway         & 30    & 30         & 1197  \\
\multicolumn{1}{c|}{11}    & Soybean-mintill               & 30    & 30         & \multicolumn{1}{c|}{2425} & Lettuce \_4wk          & 30    & 30         & \multicolumn{1}{c|}{1038}  & Railway         & 30    & 30         & 1205  \\
\multicolumn{1}{c|}{12}    & Soybean-clean                 & 30    & 30         & \multicolumn{1}{c|}{563}  & Lettuce \_5wk          & 30    & 30         & \multicolumn{1}{c|}{1897}  & Parking Lot 1   & 30    & 30         & 1203  \\
\multicolumn{1}{c|}{13}    & Wheat                         & 30    & 30         & \multicolumn{1}{c|}{175}  & Lettuce \_6wk          & 30    & 30         & \multicolumn{1}{c|}{886}   & Parking Lot 2   & 30    & 30         & 439   \\
\multicolumn{1}{c|}{14}    & Woods                         & 30    & 30         & \multicolumn{1}{c|}{1235} & Lettuce \_7wk          & 30    & 30         & \multicolumn{1}{c|}{1040}  & Tennis Court    & 30    & 30         & 398   \\
\multicolumn{1}{c|}{15}    & Buildings-grass-trees-drivers & 30    & 30         & \multicolumn{1}{c|}{356}  & Vinyard\_untrained     & 30    & 30         & \multicolumn{1}{c|}{7238}  & Running Track   & 30    & 30         & 630   \\
\multicolumn{1}{c|}{16}    & Stone-steel-towers            & 30    & 30         & \multicolumn{1}{c|}{63}   & Vinyard \_trellis      & 30    & 30         & \multicolumn{1}{c|}{1777}  &                 &       &            &       \\ \midrule
Total   &                               & 435   & 435        & 9814                      &                        & 480   & 480        & 53649                      &                 & 450   & 450        & 14579 \\ \bottomrule[1.5pt]
\end{tabular}

}
\label{tab:1}
\end{table*}

\section{Experiment}

\begin{figure}
\centerline{\includegraphics[width=0.48\textwidth]{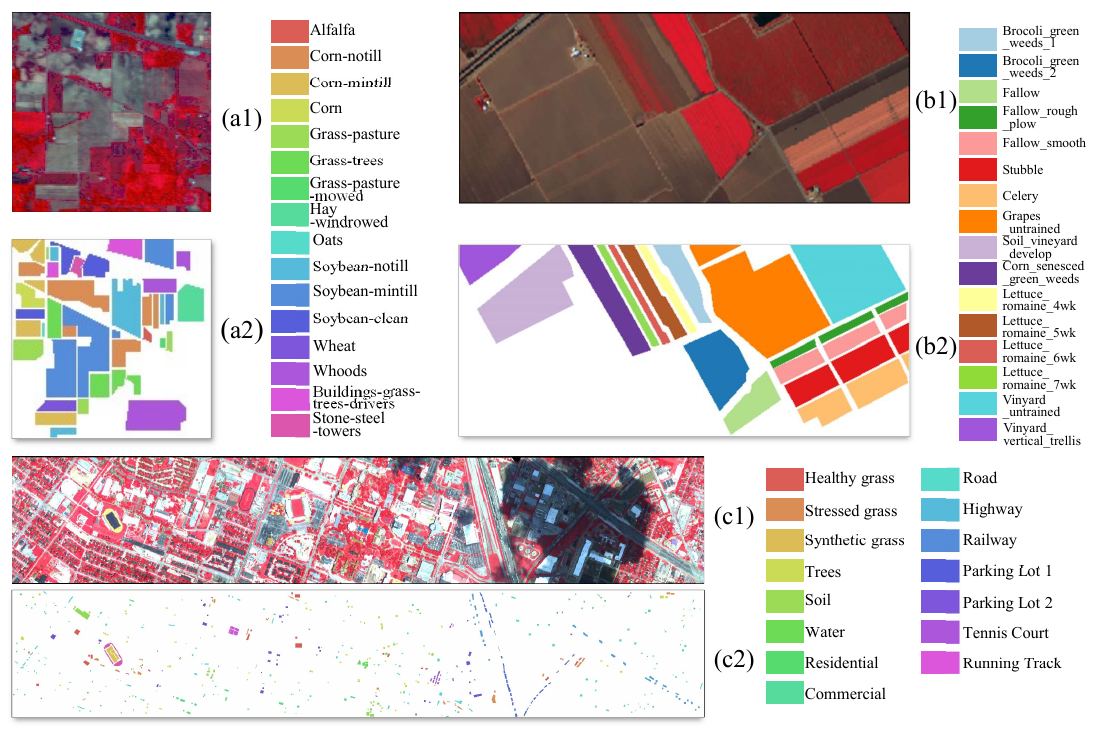}} 
\caption{Three datasets. (a1), (b1), and (c1) are the color maps of IP, SA, UH2013. (a2), (b2), and (c2) are the ground-truth maps.}
\label{fig:3}
\end{figure}

In this section, we conducted comprehensive experiments to validate the effectiveness of the proposed GraphMamba. Firstly, we evaluated the algorithm's performance by testing ten algorithms on three datasets of different sizes using three quantitative evaluation metrics. Additionally, we generated classification maps to qualitatively compare the classification results. Subsequently, we compared the advancement of the algorithms at different sampling ratios on OA and conducted a complexity analysis of the models. Finally, we verified the efficacy of the proposed modules through ablation experiments.

\begin{table*}
\caption{Quantitative comparison of IP dataset}
\resizebox{1\linewidth}{!}{  
\begin{tabular}{c|cccccccccc}
\bottomrule[1.5pt]
Class No. & SVM-RBF     & 3D CNN      & DFFN        & AB-LSTM     & RSSAN       & WFCG        & AMGCFN               & SF          & GAHT        & GraphMamba           \\ \hline
1         & 31.36±3.65  & 87.01±0.00  & 96.77±2.63  & 41.94±34.24 & 31.18±25.85 & 82.33±3.50  & 100.00±0.00          & 47.31±13.52 & 98.92±1.52  & \textbf{100.00±0.00} \\
2         & 58.16±4.99  & 41.18±6.81  & 72.22±2.52  & 41.20±15.40 & 60.94±13.57 & 85.99±3.11  & 83.08±10.17          & 50.74±5.01  & 80.31±2.26  & \textbf{94.25±1.09}  \\
3         & 51.12±5.52  & 46.92±8.72  & 91.46±3.72  & 12.62±15.86 & 64.75±9.29  & 79.80±2.77  & 88.87±7.21           & 57.04±6.71  & 84.83±7.30  & \textbf{94.41±0.83}  \\
4         & 30.90±1.75  & 88.89±2.40  & 97.58±1.18  & 41.55±27.07 & 90.02±4.96  & 75.29±2.59  & \textbf{100.00±0.00} & 84.06±6.35  & 99.03±1.04  & 98.39±2.01           \\
5         & 83.18±0.69  & 73.36±6.93  & 94.11±1.38  & 76.97±2.10  & 83.81±5.89  & 99.34±0.76  & \textbf{95.12±5.01}  & 72.33±6.60  & 91.69±3.97  & 94.63±1.85           \\
6         & 91.13±1.14  & 88.29±4.15  & 96.14±2.02  & 70.00±23.64 & 90.95±2.90  & 98.68±0.73  & 97.37±1.33           & 82.29±0.91  & 96.62±1.35  & \textbf{99.71±0.29}  \\
7         & 44.18±10.37 & 100.00±0.00 & 100.00±0.00 & 43.59±32.23 & 100.00±0.00 & 72.02±17.02 & 100.00±0.00          & 100.00±0.00 & 100.00±0.00 & \textbf{100.00±0.00} \\
8         & 95.79±1.35  & 97.02±1.00  & 97.69±0.69  & 93.01±4.86  & 99.26±0.59  & 99.48±0.27  & 99.77±0.18           & 96.35±1.24  & 99.85±0.21  & \textbf{100.00±0.00} \\
9         & 18.17±1.37  & 100.00±0.00 & 100.00±0.00 & 60.00±43.20 & 100.00±0.00 & 66.13±7.48  & 100.00±0.00          & 100.00±0.00 & 100.00±0.00 & \textbf{100.00±0.00} \\
10        & 56.76±2.69  & 57.93±3.73  & 88.46±0.30  & 51.20±18.09 & 78.91±8.61  & 69.28±4.05  & 83.24±6.53           & 71.09±4.68  & 86.13±3.30  & \textbf{95.33±0.48}  \\
11        & 73.59±3.41  & 59.64±3.73  & 68.14±1.90  & 52.04±14.98 & 56.78±8.61  & 93.44±2.42  & 90.03±6.39           & 53.46±9.35  & 75.77±4.82  & \textbf{95.30±1.14}  \\
12        & 46.22±5.87  & 62.94±5.66  & 87.63±1.48  & 41.44±2.36  & 65.78±4.34  & 82.08±8.35  & 90.98±5.12           & 43.64±4.48  & 88.93±2.04  & \textbf{93.31±2.35}  \\
13        & 86.97±0.51  & 99.05±0.27  & 99.62±0.54  & 97.71±0.47  & 98.48±0.97  & 98.89±1.56  & 99.61±0.53           & 97.71±1.62  & 99.43±0.81  & \textbf{100.00±0.00} \\
14        & 94.43±0.82  & 83.13±0.60  & 92.52±0.28  & 90.58±3.19  & 96.06±0.42  & 97.64±0.65  & 97.18±0.47           & 76.60±5.15  & 90.31±0.83  & \textbf{100.00±0.00} \\
15        & 51.70±2.47  & 80.43±6.91  & 89.51±3.78  & 54.21±11.81 & 79.78±7.24  & 82.96±5.19  & 98.77±0.81           & 85.49±3.32  & 95.51±3.39  & \textbf{99.91±0.16}  \\
16        & 83.17±8.52  & 100.00±0.00 & 99.47±0.75  & 98.41±2.24  & 98.41±1.30  & 94.53±2.37  & 99.47±0.75           & 98.94±1.50  & 100.00±0.00 & \textbf{100.00±0.00} \\ \hline
OA(\%)    & 66.96±2.25  & 65.83±1.78  & 83.61±1.01  & 56.68±3.48  & 73.78±1.04  & 87.68±0.24  & 91.25±4.84           & 65.24±2.09  & 85.96±1.69  & \textbf{96.43±0.25}  \\
AA(\%)    & 62.30±1.66  & 79.12±1.21  & 91.96±0.70  & 60.41±2.84  & 80.94±1.22  & 86.12±1.40  & 95.21±5.47           & 76.07±0.95  & 92.96±0.79  & \textbf{97.83±0.29}  \\
Kappa     & 62.81±2.42  & 61.38±1.99  & 81.47±1.15  & 51.02±3.55  & 70.48±0.99  & 86.03±0.26  & 90.03±5.47           & 61.06±2.10  & 84.05±1.88  & \textbf{95.91±0.27}  \\ \toprule[1.5pt]
\end{tabular}}
\label{tab:2}
\end{table*}

\subsection{Data Description}

Three real benchmark datasets, Indian Pines (IP), Salina (SA), and University of Houston 2013 (UH2013) are employed to test the algorithm performance from various perspectives. The IP dataset is used to evaluate the classification ability under conditions of uneven sample distribution, the SA dataset is utilized to validate the classification ability under similar spectral conditions of different land covers, and the UH2013 dataset is employed to verify the classification performance of large-scale datasets with high spatial resolution.

The Indian Pines (IP) dataset has a size of 145×145, 200 effective bands, 16 land cover classes, and a spatial resolution of approximately 20 meters.
The Salina (SA) dataset has dimensions of 512×217, 204 effective bands, 16 land cover classes, and a spatial resolution of 3.7 meters.
The University of Houston 2013 dataset is sized at 349×1905, with 144 effective bands, 15 land cover classes, and a spatial resolution of 2.5 meters.
Details of the dataset's false-color images and ground-truth maps are described in Fig.\!~\ref{fig:3}. Sample selection details of the datasets are presented in Tab.\!~\ref{fig:1}.

\subsection{Experiments Settings}\label{Settings}

\subsubsection*{\bf Comparative Methods}
To ensure the reliability of the experimental results to the fullest extent, we conducted a detailed comparison using nine of the most recent and advanced baseline experiments. These nine methods can be roughly categorized into traditional machine learning methods such as SVM-RBF~\cite{SVM-RBF}, CNN-based methods including 3D CNN~\cite{3DCNN}, DFFN~\cite{DFFN}, and RSSAN~\cite{RSSAN}, LSTM-based method AB-LSTM~\cite{AB-LSTM}, Transformer-based method SpectralFormer (SF)~\cite{SF} and GAHT~\cite{GAHT}, and fusion methods combining GCN and CNN such as WFCG~\cite{WFCG} and AMGCFN~\cite{AMGCFN}.

\begin{table*}
\caption{Quantitative comparison of SA dataset}
\resizebox{1\linewidth}{!}{  
\begin{tabular}{c|cccccccccc}
\bottomrule[1.5pt]
Class No. & SVM-RBF    & 3D CNN     & DFFN        & AB-LSTM     & RSSAN      & WFCG                 & AMGCFN              & SF         & GAHT                 & GraphMamba           \\ \hline
1         & 98.70±1.05 & 98.77±0.80 & 94.81±4.37  & 98.50±1.33  & 98.84±0.25 & 99.74±0.35           & 100.00±0.00         & 94.74±0.93 & 100.00±0.00          & \textbf{100.00±0.00} \\
2         & 99.20±0.55 & 98.70±0.71 & 99.25±0.12  & 97.06±3.51  & 99.50±0.06 & 100.00±0.00          & 99.72±0.48          & 98.82±0.49 & 99.98±0.03           & \textbf{100.00±0.00} \\
3         & 92.85±0.61 & 90.00±1.19 & 95.89±2.40  & 73.14±16.50 & 97.14±1.82 & \textbf{100.00±0.00} & 99.95±0.09          & 92.17±1.99 & 98.49±0.96           & 99.90±0.18           \\
4         & 97.54±0.17 & 98.75±0.36 & 96.51±1.00  & 99.49±0.27  & 99.49±0.41 & 98.99±0.38           & 98.53±2.23          & 96.26±0.94 & 99.12±0.75           & \textbf{99.73±0.35}  \\
5         & 97.65±0.65 & 95.20±1.07 & 97.52±0.69  & 93.76±2.99  & 96.99±1.21 & 97.87±0.71           & 97.25±0.73          & 89.41±1.83 & \textbf{99.13±0.44}  & 95.90±0.46           \\
6         & 99.89±0.02 & 99.40±0.39 & 99.47±0.73  & 96.77±2.37  & 99.02±0.66 & 99.96±0.04           & 99.59±0.63          & 99.20±0.72 & \textbf{99.97±0.04}  & 99.49±0.44           \\
7         & 98.76±0.43 & 99.38±0.37 & 99.49±0.30  & 98.96±0.38  & 98.07±1.66 & 99.80±0.27           & \textbf{99.99±0.02} & 97.91±1.99 & 99.80±0.16           & 99.56±0.33           \\
8         & 77.68±1.55 & 72.55±6.82 & 79.11±2.26  & 41.24±30.46 & 71.60±1.74 & 77.52±7.72           & 83.51±9.23          & 73.07±1.12 & 83.11±3.18           & \textbf{91.64±1.02}  \\
9         & 99.18±0.24 & 93.30±0.93 & 94.76±0.58  & 91.23±6.09  & 98.15±0.96 & 99.90±0.11           & 99.96±0.07          & 93.77±1.42 & 99.47±0.11           & \textbf{100.00±0.00} \\
10        & 83.99±1.33 & 88.24±0.95 & 91.81±0.85  & 47.84±23.55 & 92.03±2.89 & 94.71±2.08           & \textbf{99.61±0.34} & 91.39±1.83 & 97.36±1.81           & 97.32±0.43           \\
11        & 89.42±0.31 & 93.96±2.26 & 98.75±0.75  & 82.69±11.92 & 96.76±1.87 & 98.74±0.56           & 99.33±0.59          & 92.36±3.31 & 98.62±1.14           & \textbf{100.00±0.00} \\
12        & 95.31±0.35 & 99.33±0.22 & 99.95±0.07  & 95.01±1.29  & 99.37±0.75 & 100.00±0.00          & 99.91±0.15          & 99.05±0.58 & \textbf{100.00±0.00} & 99.95±0.00           \\
13        & 94.17±1.21 & 99.06±0.93 & 100.00±0.00 & 86.61±15.19 & 99.44±0.65 & 99.96±0.05           & 99.85±0.13          & 99.40±0.43 & 99.96±0.05           & \textbf{100.00±0.00} \\
14        & 90.11±4.68 & 96.73±0.95 & 99.26±0.32  & 92.98±2.45  & 95.54±2.70 & 98.04±0.76           & 99.23±0.39          & 96.60±0.95 & 98.94±0.21           & \textbf{99.90±0.00}  \\
15        & 60.59±1.96 & 71.35±3.32 & 83.37±0.37  & 58.63±30.10 & 79.07±3.13 & 84.93±7.15           & 89.77±10.32         & 81.58±2.20 & 87.10±2.83           & \textbf{96.88±0.35}  \\
16        & 97.04±1.20 & 92.61±1.97 & 93.60±1.91  & 88.37±8.03  & 95.72±2.55 & 99.15±0.45           & \textbf{99.90±0.17} & 93.72±2.04 & 96.83±2.66           & 98.31±0.93           \\ \hline
OA(\%)    & 87.31±0.53 & 87.59±1.49 & 91.35±0.41  & 74.83±1.60  & 89.66±0.76 & 92.68±1.07           & 94.87±1.07          & 88.80±0.21 & 94.21±0.39           & \textbf{97.33±0.07}  \\ 
AA(\%)    & 92.01±0.15 & 92.96±0.56 & 95.22±0.13  & 83.89±3.61  & 94.79±0.71 & 96.83±0.40           & 97.88±0.50          & 93.09±0.19 & 97.37±0.07           & \textbf{98.66±0.12}  \\
Kappa     & 85.90±0.58 & 86.21±1.64 & 90.39±0.45  & 72.21±1.69  & 88.53±0.83 & 91.86±1.18           & 94.30±1.19          & 87.57±0.24 & 93.56±0.43           & \textbf{97.02±0.07}  \\ \toprule[1.5pt]
\end{tabular}}
\label{tab:3}
\end{table*}

\begin{table*}[!h]
\caption{Quantitative comparison of UH2013 dataset}
\resizebox{1\linewidth}{!}{  
\begin{tabular}{c|cccccccccc}
\bottomrule[1.5pt]
Class No. & SVM-RBF    & 3D CNN     & DFFN                & AB-LSTM    & RSSAN       & WFCG        & AMGCFN      & SF         & GAHT                & GraphMamba           \\ \hline
1         & 90.95±0.01 & 94.02±6.56 & 94.43±6.09          & 89.90±4.83 & 90.72±4.49  & 90.39±5.30  & 86.09±1.37  & 91.10±3.81 & 95.55±2.35          & \textbf{96.64±1.04}  \\
2         & 94.80±0.02 & 81.62±8.94 & 96.81±4.28          & 93.33±1.96 & 97.49±1.55  & 94.74±6.26  & 87.84±9.21  & 93.19±2.59 & \textbf{98.88±0.27} & 98.86±0.17           \\
3         & 98.61±0.01 & 96.8±1.10  & 99.95±0.07          & 98.55±0.63 & 99.15±0.67  & 99.97±0.05  & 99.35±0.25  & 93.90±4.60 & 100.00±0.00         & \textbf{100.00±0.00} \\
4         & 98.73±0.01 & 94.15±3.08 & 94.54±3.20          & 93.93±2.49 & 95.09±3.42  & 96.84±2.52  & 90.04±1.91  & 88.44±2.62 & 94.23±1.04          & \textbf{100.00±0.00} \\
5         & 92.38±0.04 & 98.79±1.11 & 98.46±1.95          & 87.35±7.38 & 94.39±5.66  & 99.83±0.03  & 99.42±0.75  & 97.14±0.69 & 99.39±0.74          & \textbf{100.00±0.00} \\
6         & 95.52±0.05 & 84.29±3.69 & 98.42±1.39          & 92.66±6.90 & 89.83±5.54  & 90.56±3.34  & 92.65±0.63  & 88.36±3.93 & 94.58±5.27          & \textbf{100.00±0.00} \\
7         & 80.98±0.03 & 77.95±5.29 & 85.73±2.51          & 68.66±3.31 & 84.41±3.82  & 94.27±1.71  & 92.16±2.55  & 74.18±1.90 & 84.3±3.73           & \textbf{94.40±1.50}  \\
8         & 78.25±0.03 & 54.37±3.16 & 79.41±7.35          & 48.00±2.54 & 71.17±4.95  & 64.18±5.78  & 71.92±3.56  & 76.44±4.59 & 82.18±4.18          & \textbf{83.52±2.11}  \\
9         & 72.56±0.01 & 81.23±1.67 & \textbf{85.00±1.35} & 75.78±1.49 & 83.42±4.80  & 81.58±4.46  & 76.28±4.50  & 67.59±3.71 & 83.77±3.51          & 83.36±0.58           \\
10        & 79.96±0.03 & 63.30±3.48 & 90.98±4.38          & 36.34±3.18 & 68.64±10.93 & 74.15±11.74 & 87.49±14.82 & 78.75±3.61 & 89.75±5.84          & \textbf{99.70±0.29}  \\
11        & 79.84±0.05 & 62.79±2.97 & 82.71±6.13          & 66.67±8.79 & 66.25±8.14  & 81.06±6.09  & 88.98±4.43  & 62.38±6.03 & 77.48±0.81          & \textbf{99.92±0.09}  \\
12        & 75.41±0.03 & 55.20±9.18 & 83.07±7.80          & 30.29±9.48 & 79.30±5.36  & 74.62±9.77  & 82.76±12.08 & 74.40±5.53 & 88.03±2.18          & \textbf{93.57±0.33}  \\
13        & 50.96±0.04 & 92.26±2.14 & \textbf{95.60±0.88} & 33.41±6.31 & 92.71±1.12  & 87.19±3.59  & 88.61±0.67  & 64.16±5.65 & 94.08±1.48          & 93.47±1.03           \\
14        & 88.51±0.06 & 98.99±0.74 & 99.50±0.54          & 96.31±0.43 & 69.93±41.63 & 99.94±0.10 & 100.00±0.00 & 88.86±1.48 & 99.75±0.21          & \textbf{100.00±0.00} \\
15        & 97.14±0.01 & 99.95±0.07 & 99.74±0.20          & 96.51±1.62 & 93.02±8.88  & 100.00±0.00 & 100.00±0.00 & 98.52±0.92 & 100.00±0.00         & \textbf{100.00±0.00} \\ \hline
OA(\%)    & 84.98±0.01 & 79.61±1.36 & 90.74±1.67          & 71.81±1.49 & 84.36±3.09  & 87.11±2.15  & 88.07±2.05  & 81.68±0.68 & 90.84±0.59          & \textbf{95.62±0.03}  \\ 
AA(\%)    & 84.97±0.02 & 82.38±1.31 & 92.29±1.31          & 73.84±1.50 & 85.03±4.64  & 88.63±1.76  & 89.58±1.67  & 82.49±0.14 & 92.13±0.64          & \textbf{96.23±0.07}  \\
Kappa     & 83.75±0.01 & 77.97±1.46 & 89.99±1.81          & 69.54±1.62 & 83.08±3.35  & 86.07±2.33  & 87.09±2.22  & 80.22±0.71 & 90.10±0.63          & \textbf{95.27±0.03}   \\ \toprule[1.5pt]
\end{tabular}}
\label{tab:4}
\end{table*}

\subsubsection*{\bf Evaluation Metrics}
The overall classification accuracy (OA) is used to observe the classification accuracy of all samples. The average accuracy (AA) is utilized to represent the average accuracy of different class classifications. The Kappa coefficient is employed as an indicator of the degree of consistency. Additionally, the classification accuracy of each class is specifically presented.

\subsubsection*{\bf Parameter Settings and Implementation Details}
For a fair comparison, all comparative methods were run on an NVIDIA Titan RTX GPU using the PyTorch framework and Python 3.8. During training, the Adam optimizer was employed with a learning rate of ${\rm{5}} \times {\rm{1}}{{\rm{0}}^{{\rm{ - 4}}}}$. Epoch=200.
For GraphMamba, we used a patch size of 11×11 and a depth of 8 for the encoding module.

\subsection{Comparative Experimental Results and Analysis}

\begin{figure*}
\centerline{\includegraphics[width=0.98\textwidth]{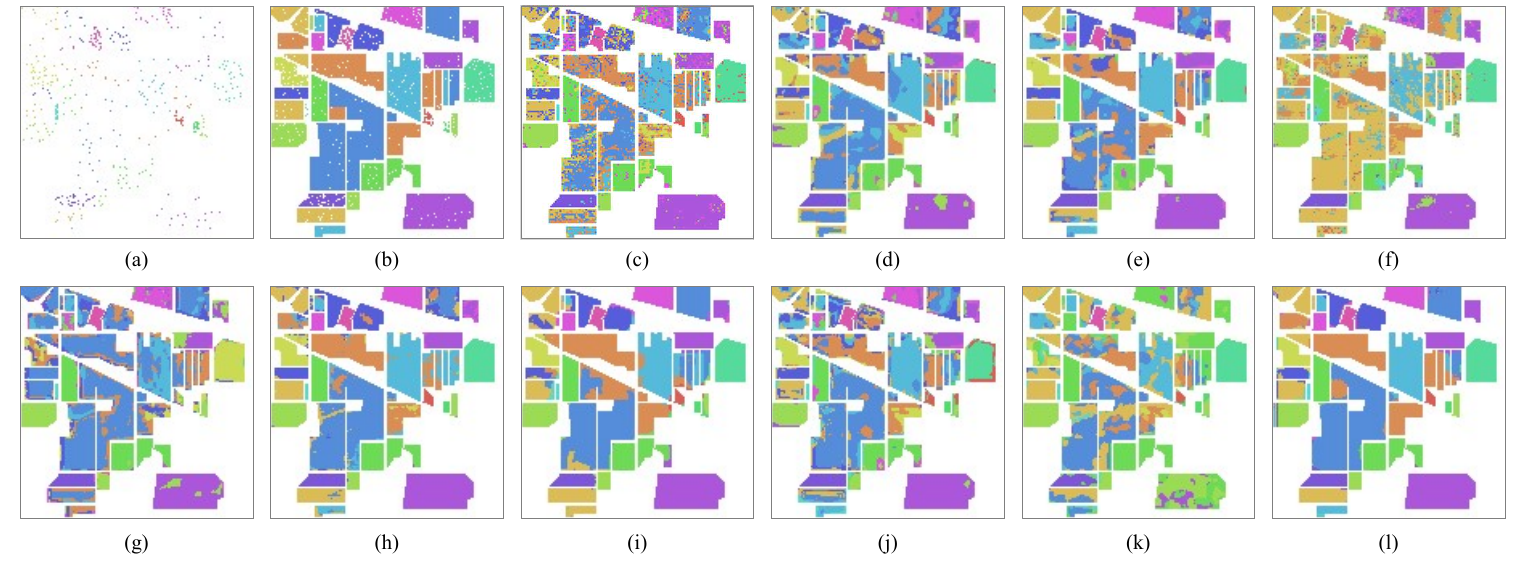}} 
\caption{Classification maps of IP. (a) Train Label (b) Test Label (c) SVM-RBF  (d) 3D CNN (e) DFFN (f) AB-LSTM (g) RSSAN (h) WFCG (i) AMGCFN (j) SF (k) GAHT (l) GraphMamba}
\label{fig:4}
\end{figure*}

\begin{figure*}
\centerline{\includegraphics[width=0.98\textwidth]{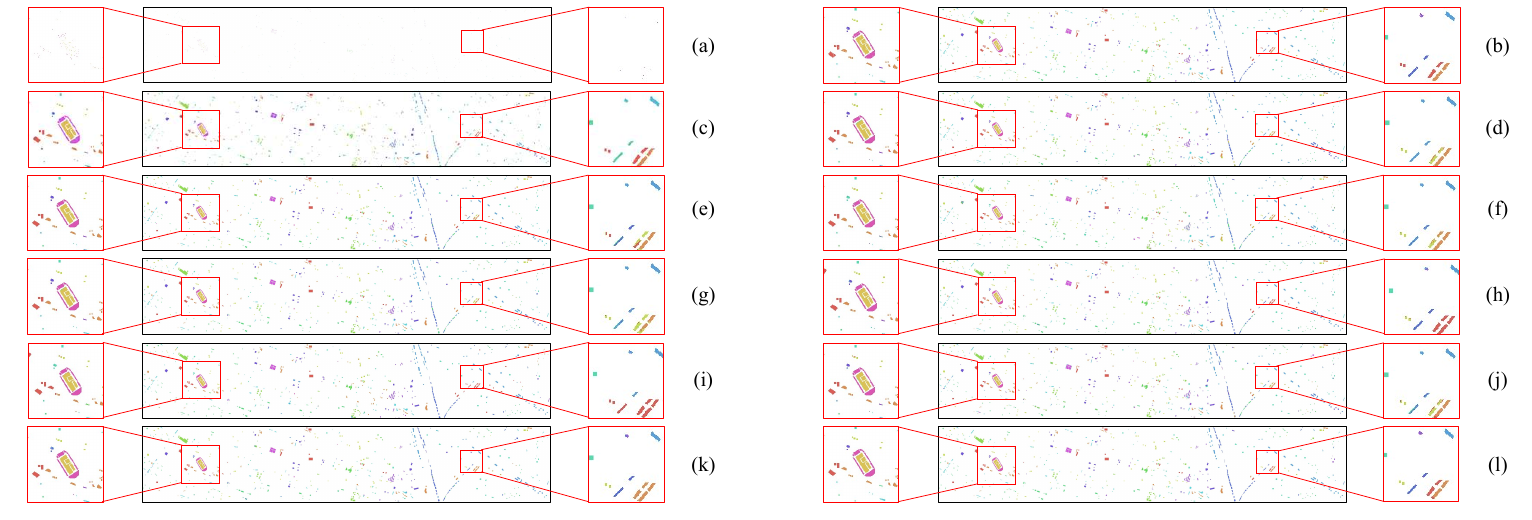}} 
\caption{Classification maps of UH2013. (a) Train Label (b) Test Label (c) SVM-RBF (d) 3D CNN (e) DFFN (f) AB-LSTM (g) RSSAN (h) WFCG (i) AMGCFN (j) SF (k) GAHT (l) GraphMamba}
\label{fig:6}
\end{figure*}

In this section, we qualitatively and quantitatively evaluate the IP, SA, and UH2013 datasets, and detail the clustering results in Tab.\!~\ref{tab:2}-\!\!~\ref{tab:4}, highlighting the best performance in bold. Additionally, the intuitive classification results are shown in Fig.\!~\ref{fig:4}-\!\!~\ref{fig:6}.

\begin{figure*}
\centerline{\includegraphics[width=0.98\textwidth]{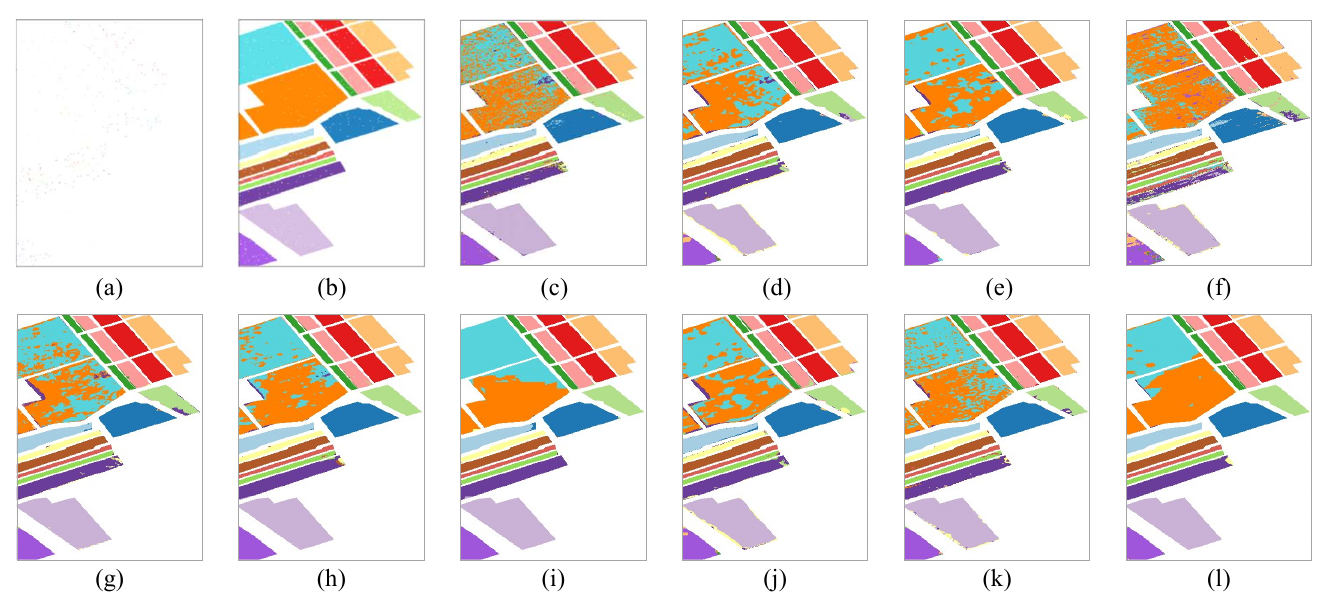}} 
\caption{Classification maps of SA. (a) Train Label (b) Test Label (c) SVM-RBF  (d) 3D CNN (e) DFFN (f) AB-LSTM (g) RSSAN (h) WFCG (i) AMGCFN (j) SF (k) GAHT (l) GraphMamba.}
\label{fig:5}
\end{figure*}

\subsubsection*{\bf Quantitative analysis of Indian Pines dataset}

Tab.\!~\ref{tab:2} displays the comparison results of ten different methods with GraphMamba, with the best classification performance highlighted in each row. We selected various baseline methods such as CNN, LSTM, GCN, Transformer, and their improved or fused versions for comparison in the experiments to comprehensively test Mamba's performance in HSI classification. From the table, we observe that the LSTM-based network has lower classification accuracy as it solely focuses on spectral aspects, neglecting spatial features. The CNN and Transformer-based methods exhibit better classification results by effectively capturing local spectral information. Methods based on GCN, such as WFCG and AMGCFN, achieve good classification results due to their strong spatial-spectral modeling capabilities.

Notably, compared to other models, GraphMamba achieves the best classification performance in terms of OA, AA, and Kappa metrics, with improvements of 5.18\%, 2.62\%, and 5.88\%, respectively, over the second-best performing model. This demonstrates the effectiveness of the proposed graph-based learning approach in Mamba.

In terms of visualization results, traditional methods like SVM-RBF tend to exhibit salt-and-pepper noise, while deep learning-based classification methods produce smoother classification maps. Furthermore, our method shows better handling of edge features, significantly reducing edge noise. This is attributed to the context-aware GCN encoding module used, which better aggregates spatial information and reduces misclassification instances. Additionally, the HyperMamba encoder enhances the modeling of sequences, reducing noise during the classification process.

\subsubsection*{\bf Quantitative analysis of Salina dataset}

The presence of similar terrain features in the SA dataset poses a challenge for classification. In this scenario, GraphorMamba continues to exhibit the most dominant classification results, with improvements of 2.46\%, 0.78\%, and 2.72\% in OA, AA, and Kappa, respectively, compared to the second-best classification method. Furthermore, the classification results of our approach have the lowest standard deviation, indicating a high level of algorithmic stability. Due to the spectral similarity between Grapes untrained and Vinyard untrained land covers, their classification accuracy is relatively low. However, GraphMamba achieves superior classification results among all methods by integrating spatial-spectral features more effectively through a multi-scale contextual attention mechanism.

From the classification maps, it is evident that GraphMamba delivers satisfactory classification results even with a small proportion of training data. Given the proximity of terrain features, misclassifications and noise are prone to occur at the edges. GraphMamba effectively addresses this issue, resulting in a smoother visual outcome.

\subsubsection*{\bf Quantitative analysis of UH2013 dataset}

For the large-scale UH2013 dataset, its vast size and widely scattered land covers pose significant challenges for precise classification. GraphMamba utilized only 450 pixels (2.99\% of the total pixels) and achieved an overall accuracy of 95.62\%. Compared to the second-best classification performance, it exhibited improvements of 4.78\% in OA, 4.1\% in AA, and 5.17\% in KAPPA. The smaller standard deviation indicates the stability of its algorithm. The inherent advantages of Mamba in HyperMamba, along with the proposed global masking module, enable better capture of spectral information. In contrast to CNN, the SpatialGCN module in HyperMamba can handle spatial information more flexibly and is equally effective in scenarios with large-scale scattered land covers.
Zooming into the enlarged area in Fig.\!~\ref{fig:6} reveals that even very small land covers can be accurately classified, whereas other methods tend to misclassify entire small land covers. Additionally, GraphMamba's handling of edges is quite satisfactory. This demonstrates that joint spatial-spectral information for feature modeling can better represent relationships between land covers, thereby enhancing classification performance.

\begin{figure*}
\centerline{\includegraphics[width=1\textwidth]{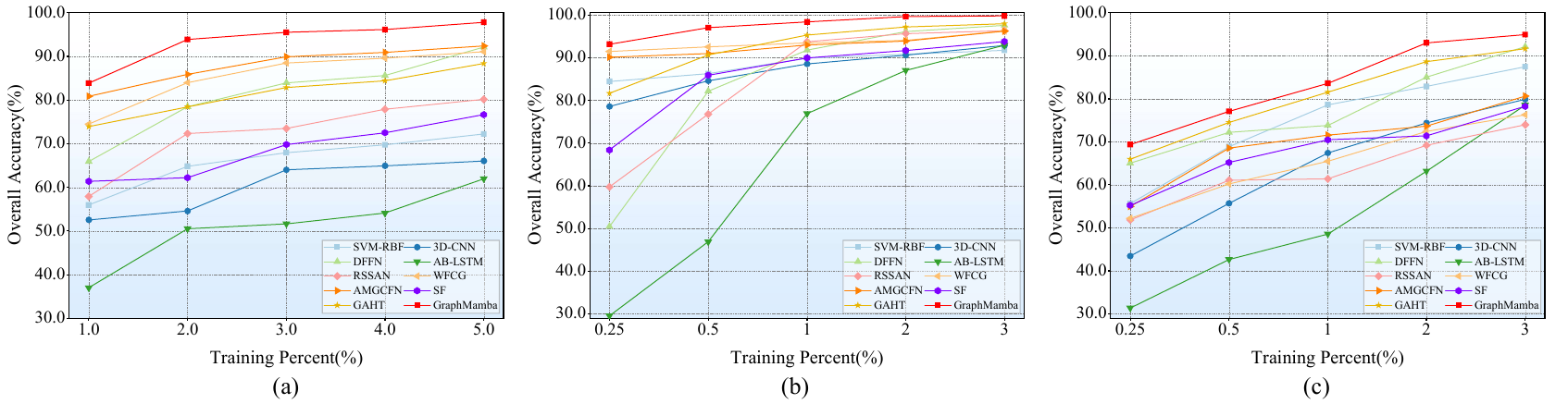}} 
\caption{Three datasets. (a1), (b1), and (c1) are the color maps of IP, SA, UH2013. (a2), (b2), and (c2) are the ground-truth maps.}
\label{fig:8}
\end{figure*}

\subsection{Analysis of Training Samples}

Different sampling ratios may lead to performance differences. To evaluate the adaptability of GraphMamba under different proportions of training samples, experiments were conducted using five different sampling ratios across ten methods. Specifically, for the IP dataset, training samples are taken at 1\%, 2\%, 3\%, 4\%, and 5\% of the dataset, while for the SA and UH2013 datasets, training samples are taken at 0.25\%, 0.5\%, 1\%, 2\%, and 3\%. Fig.\!~\ref{fig:8} illustrates how the OA varies with the number of samples on the three datasets. As the number of samples increases, the OA also increases. It is noteworthy that the OA of GraphMamba (as indicated by the red line) consistently remains at the highest level, demonstrating its strong robustness and generalization capability. This is attributed to GraphMamba's adaptive spatial contextual modeling and spectral feature extraction capabilities.

\begin{figure}
\centerline{\includegraphics[width=0.5\textwidth]{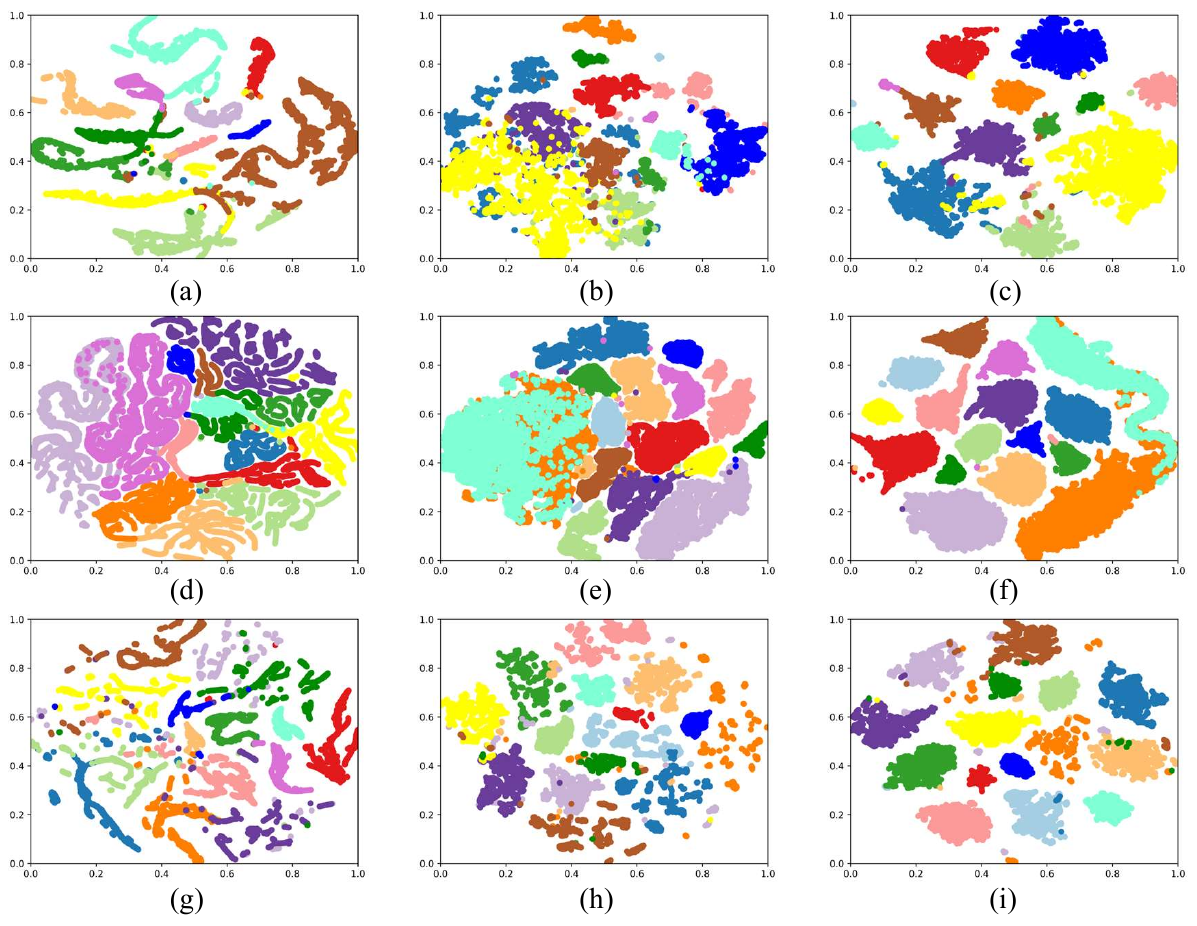}} 
\caption{Feature separability across various methods in three datasets: (a) AMGCFN, (b) GAHT, and (c) GraphMamba for IP. (d) AMGCFN, (e) GAHT, and (f) GraphMamba for SA.(g) AMGCFN, (h) GAHT, and (i) GraphMamba for UH2013}
\label{fig:7}
\end{figure}

\subsection{Visualization Analysis of Features}

Fig.\!~\ref{fig:7} shows the feature separation graphs of the classification features of AMGCFN, GAHT, and GraphMamba after PCA dimensionality reduction on three datasets, respectively. It can be observed that the features of GraphMamba are more aggregated, with larger distances between features of different classes. In contrast, AMGCFN and GAHT exhibit more overlapping and intersecting parts between features of different classes, with features of the same class being more scattered. This indicates that GraphMamba has a strong capability in spectral information mining, enabling the extraction of more discriminative information. This also suggests a broad application prospect for Mamba in HSI classification.

\begin{figure*}
\centerline{\includegraphics[width=1\textwidth]{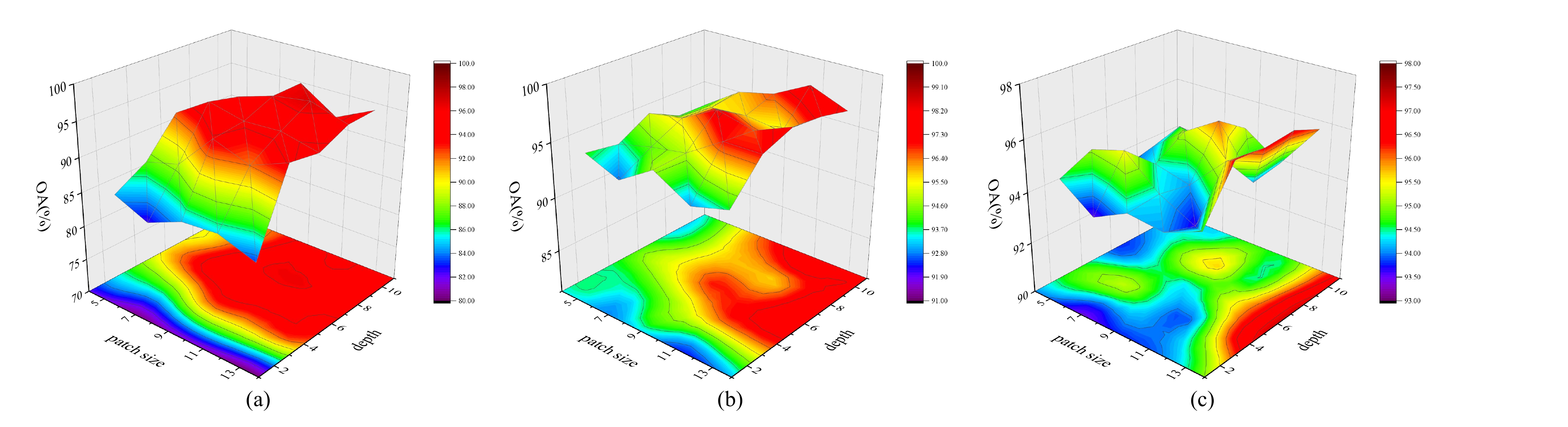}} 
\caption{Sensitivity of GraphMamba in different parameter settings(i.e., patch size and encoder depth) on (a) IP, (b) Salinas, and (c) UH2013}
\label{fig:9}
\end{figure*}

\begin{table}[t]
\caption{OA, AA, and KAPPA coefficients of IP dataset under different module settings}
\scriptsize
\resizebox{1\linewidth}{!}{
\begin{tabular}{c|cccc}
\bottomrule[1.5pt]
Method & GraphMamba & GM-A  & GM-B  & GM-C  \\ \hline
OA     & \textbf{96.71}    & 95.53 & 93.38 & 93.88 \\
AA     & \textbf{97.98}    & 94.29 & 96.86 & 93.23 \\
Kappa  & \textbf{96.23}    & 94.90 & 92.45 & 93.00 \\ \toprule[1.5pt]
\end{tabular}}
\label{tab:5}
\end{table}

\begin{table}[t]
\caption{OA, AA, and KAPPA coefficients of SA dataset under different module settings}
\scriptsize
\resizebox{1\linewidth}{!}{
\begin{tabular}{c|cccc}
\bottomrule[1.5pt]
Method & GraphMamba & GM-A  & GM-B  & GM-C  \\ \hline
OA     & \textbf{97.36}    & 95.17 & 94.34 & 95.54 \\
AA     & \textbf{98.60}    & 97.89 & 97.67 & 97.88 \\
Kappa  & \textbf{97.06}    & 94.61 & 93.69 & 95.03 \\ \toprule[1.5pt]
\end{tabular}}
\label{tab:6}
\end{table}

\begin{table}[!t]
\caption{OA, AA, and KAPPA coefficients of UH2013 dataset under different module settings}
\scriptsize
\resizebox{1\linewidth}{!}{
\begin{tabular}{c|cccc}
\bottomrule[1.5pt]
Method & GraphMamba     & GM-A  & GM-B  & GM-C  \\ \hline
OA     & \textbf{95.54} & 94.11 & 93.81 & 94.20 \\
AA     & \textbf{96.17} & 95.18 & 94.81 & 95.18 \\
Kappa  & \textbf{95.18} & 93.63 & 93.31 & 93.73 \\ \toprule[1.5pt]
\end{tabular}}
\label{tab:7}
\end{table}

\subsection{ Ablation Experiment}

To validate the effectiveness of different mechanisms, we designed ablation experiments in this section. GM-A represents the removal of the Global Mask, GM-B represents the exclusion of the adaptive skip-connection mechanism, and GM-C represents the elimination of the adaptive filtering mechanism, with the remaining experimental settings consistent with Sec.\!~\ref{Settings}. The classification results of these methods on three datasets are shown in Tab.\!~\ref{tab:5}-\!\!~\ref{tab:7}. It can be observed that removing these modules leads to a decrease in model performance, indicating that the global mask, adaptive skip-connection, and adaptive filtering modules all make significant contributions to improving classification accuracy by better jointly extracting spatial-spectral structural information for precise classification.

\subsection{Hyperparameter Impact}

In this section, we investigated the influence of different hyperparameters on the classification performance of the network. We employed a grid search strategy to find the optimal settings for Patch size and encoder quantity. As shown in Fig.\!~\ref{fig:9}, we provided the OA for two different parameter values on three datasets.

As shown in Fig.\!~\ref{fig:9}, the overall classification performance is better when Patch=11. Moreover, the classification results do not improve by blindly increasing the patch size, as too few land object details make it difficult to capture spatial features, while overly large patches are prone to mixing irrelevant land object information. Regarding the number of encoders, the overall performance is best when N=8. Excessive network depth greatly consumes computational resources, making training more challenging. Therefore, choosing an appropriate number of encoders can extract spatial-spectral features effectively while meeting practical requirements.

\begin{table}
\caption{Comparison of running efficiency for different methods}
\resizebox{1\linewidth}{!}{
\begin{tabular}{ccccccc|c}
\bottomrule[1.5pt]
\multirow{2}{*}{Methods} & \multicolumn{6}{c|}{Running Time (s)}                                                               & \multirow{3}{*}{\begin{tabular}[c]{@{}c@{}}Complexity\\    (FLOPs)\end{tabular}} \\ \cline{2-7}
                        & \multicolumn{2}{c}{IP}          & \multicolumn{2}{c}{SA}           & \multicolumn{2}{c|}{UH2013}    &                                                                                  \\ \cline{1-7}
                        & Train           & Test          & Train           & Test           & Train          & Test          &                                                                                  \\
3D CNN                  & 76.27           & 3.16          & 81.60           & 16.92          & 65.34          & 78.01         & 0.69G                                                                            \\
DFFN                    & 75.68           & 2.09          & 78.30           & 10.99          & 38.46          & 60.84         & 4.27G                                                                            \\
RSSAN                   & 57.00           & 2.08          & 61.06           & 9.10           & 30.98          & 51.63         & 0.85G                                                                            \\
SF                      & 470.37          & 46.05         & 510.01          & 254.53         & 451.55         & 2880.05       & 1.50G                                                                            \\
GAHT                    & 75.26           & 2.91          & 81.18           & 15.07          & 47.51          & 89.80         & 4.72G                                                                            \\
AMGCFN                  & 83.54           & 4.99          & 142.24          & 7.13           & 132.56         & 8.89          & 6.32G                                                                            \\
WFCG                    & 204.45          & 1.61          & 845.55          & 7.04           & 621.15         & 11.32         & 5.98G                                                                            \\
\textbf{GraphMamba}     & \textbf{120.56} & \textbf{4.65} & \textbf{200.89} & \textbf{11.82} & \textbf{23.41} & \textbf{5.68} & \textbf{0.69G}                                                                   \\ \toprule[1.5pt]
\end{tabular}}
\label{tab:8}
\end{table}

\subsection{Comparison of runtime and model complexity}

In this section, we compare the runtime and model complexity of all algorithms. Tab.\!~\ref{tab:8} presents the training and testing times(s) of GraphMamba on three datasets, along with model complexity characterized by FLOPs. It can be observed that the runtime of GraphMamba is within an acceptable range. It is noteworthy that HyperMamba employs efficient parallel computing to reduce time, greatly enhancing computational efficiency. Additionally, GraphMamba exhibits lower complexity. Therefore, the combination of superior classification performance and lower training costs positions GraphMamba with broad prospects for industrial applications.

\section{Conclusion}

Spectral redundancy, complex spatial relationships, and the issue of model computational resource consumption have long been focal points in the field of hyperspectral image classification. This paper addresses these challenges by introducing an efficient graph structure learning vision Mamba for HSI classification, called GraphMamba. GraphMamba presents a universal paradigm for processing hyperspectral features, facilitating the extraction of discriminative features more conveniently and efficiently. Furthermore, we have designed the HyperMamba module to eliminate spectral redundancy, and the parallel computing approach significantly enhances computational efficiency and reduces overhead costs. SpatialGCN leverages adaptive context awareness and spatial filtering matrices to explore spatial features. Experimental results on three real datasets of varying scales demonstrate that, compared to various baseline methods, the proposed GraphMamba achieves more precise classification through the integrated use of spatial-spectral information, exhibiting superior performance.

{
\bibliographystyle{IEEEtran}
\bibliography{IEEEabrv,egbib}
}
\vspace{-40pt}
\begin{IEEEbiography}
[{\includegraphics[width=1in,height=1.25in,clip,keepaspectratio]{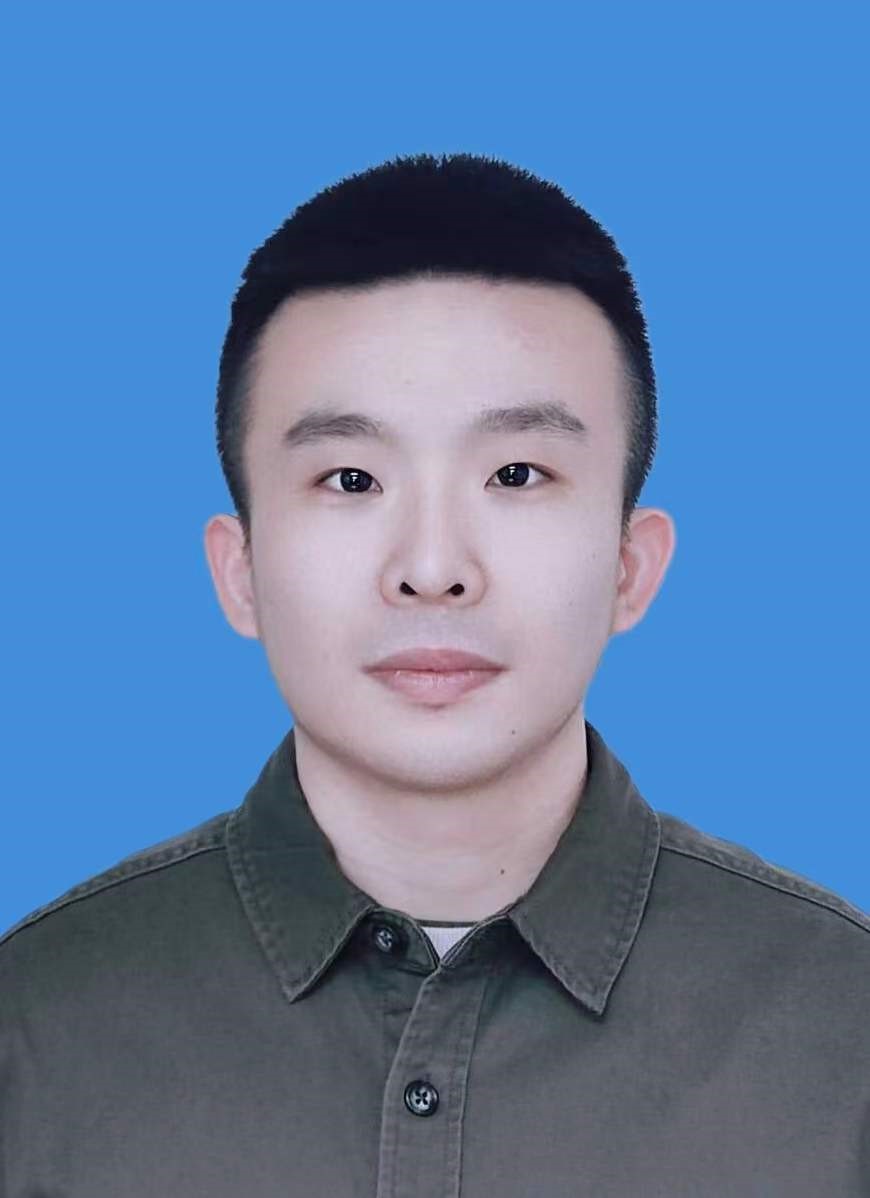}}]{Aitao Yang}
received the B.S. degree in information engineering from Xi'an Institute of High Technology, Xi'an, China, in 2017. He is currently pursuing the Ph.D. degree in computer science and technology at Xi’an Hi-Tech Research Institute, in 2021.

His main research interests include remote sensing information processing, hyperspectral image clustering, and computer vision.
\end{IEEEbiography}
\vspace{-25pt}

\begin{IEEEbiography}
[{\includegraphics[width=1in,height=1.25in,clip,keepaspectratio]{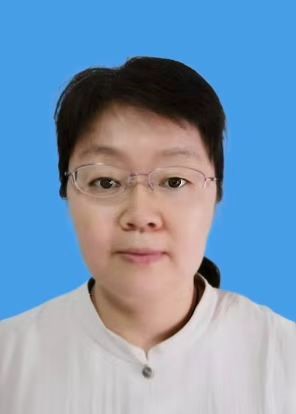}}]{Min Li}
received the B.S. and M.S. degree in computer science and technology from Xi'an Institute of High Technology, Xi'an, China, in 1991 and 1994, respectively,adn the Ph.D.degree in communication and information from Xi’an Jiaotong University, Xi'an, in 1998.

She is currently a Professor with the Xi'an Institute of High Technology. Her research interests include remote sensing, synthetic aperture radar (SAR) image processing, target detection and recognition, and feature extraction.

\end{IEEEbiography}
\vspace{-30pt}

\begin{IEEEbiography}
[{\includegraphics[width=0.9in,height=1.25in,clip,keepaspectratio]{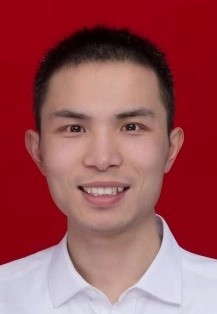}}]{Yao Ding}
received the M.S. and Ph.D. degree from the Key Laboratory of Optical Engineering, Xi’an Research Institute of High Technology, Xi’an 710025, China, in 2013 and 2022. His research interests include neural network, computer vision, image processing, and hyperspectral image clustering. He has published several papers in IEEE Trans. on Geoscience and Remote Sensing (TGRS), Information Sciences (INS), Expert Systems with Applications (ESWA), Defence Technology (DT), IEEE Geoscience and Remote sensing Letters (GRSL), Neurocomputing, etc. Furthermore, he has published three monographs, and six patents have been applied. He has received excellent doctoral dissertations from the China Simulation Society and the China Ordnance Industry Society in 2023. He also has received HIGHLY CITED AWARDS from Defence Technology (DT) journal. At present, he has Ten highly cited papers of ESI. In addition, he is also the reviewer of TGRS, TNNLS, PR, JAG, KBS. He has also served as a Guest Editor of the Forecasting.
\end{IEEEbiography}
\vspace{-30pt}

\begin{IEEEbiography}
[{\includegraphics[width=1in,height=1.25in,clip,keepaspectratio]{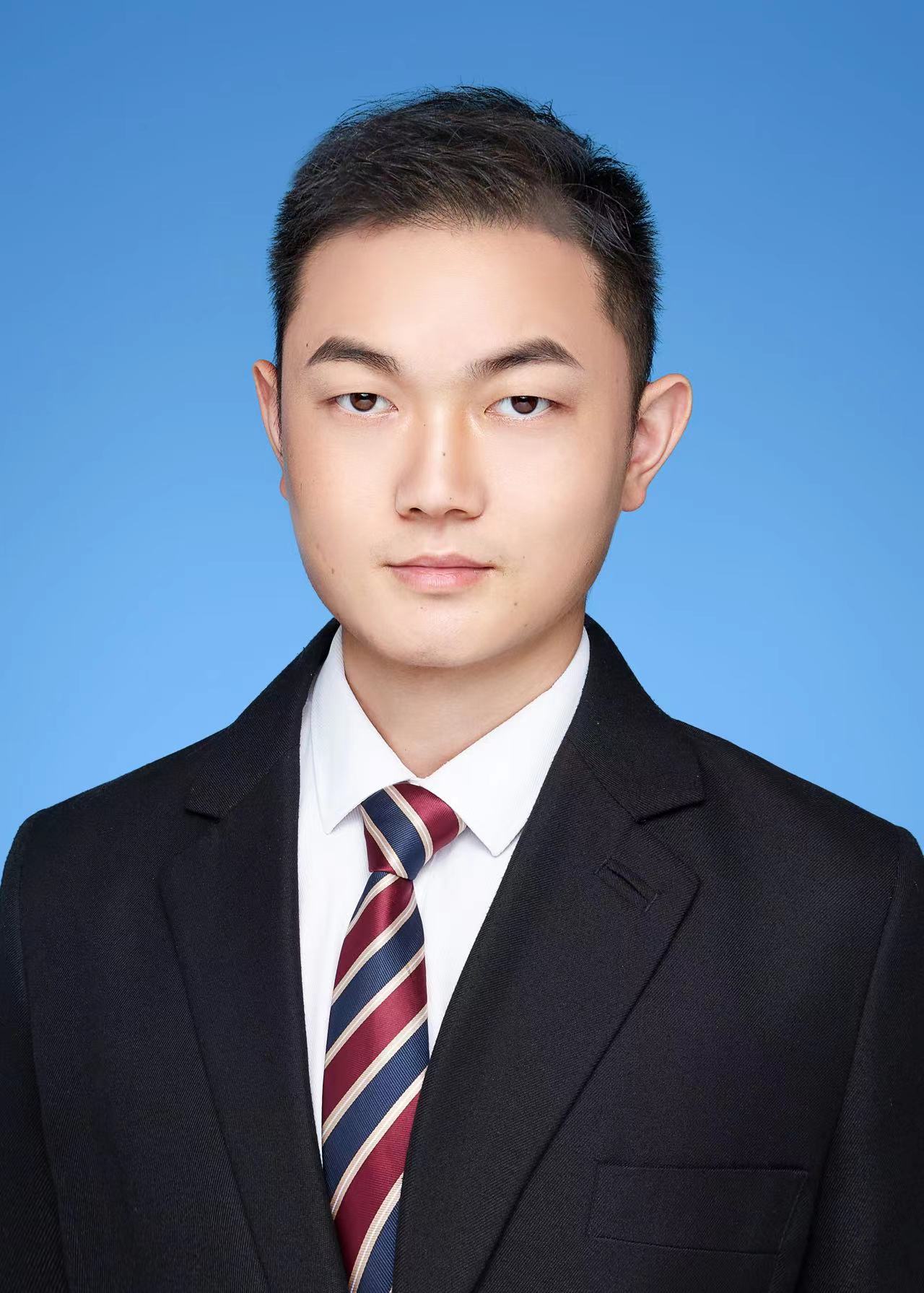}}]{Leyuan Fang}
(Senior Member, IEEE) received the Ph.D. degree from the College of Electrical and Information Engineering, Hunan University, Changsha, China, in 2015. 

From August 2016 to September 2017, he was a Post-Doctoral Researcher with the Department of Biomedical Engineering, Duke University, Durham, NC, USA. He is currently a Professor with the College of Electrical and Information Engineering, Hunan University. His research interests include sparse representation and multiresolution analysis in remote sensing and medical image processing. 

Dr. Fang was a recipient of one Second-Grade National Award at the Nature and Science Progress of China in 2019. He is an Associate Editor of IEEE TRANSACTIONS ON IMAGE PROCESSING, IEEE TRANSACTIONS ON GEOSCIENCE AND REMOTE SENSING, IEEE TRANSACTIONS ON NEURAL NETWORKS AND LEARNING SYSTEMS, and Neurocomputing.
\end{IEEEbiography}
\vspace{-30pt}

\begin{IEEEbiography}
[{\includegraphics[width=1in,height=1.25in,clip,keepaspectratio]{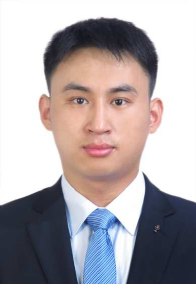}}]{Yaoming Cai}
(Member, IEEE) received the B.E. degree in Information Security and Ph.D. degree in Geoinformation Engineering from China University of Geosciences, Wuhan, China, in 2016 and 2022, respectively.  He was a Visiting Scholar with Helmholtz-Zentrum Dresden-Rossendorf, Helmholtz Institute Freiberg for Resource Technology, Freiberg, Germany, from 2021 to 2022. He is currently an Associate Professor at the School of Information Engineering, Zhongnan University of Economics and Law, Wuhan, China. His research focuses on machine learning, multimodal learning, intelligent optimization, and remote image processing.
\end{IEEEbiography}
\vspace{-30pt}

\begin{IEEEbiography}
[{\includegraphics[width=1in,height=1.25in,clip,keepaspectratio]{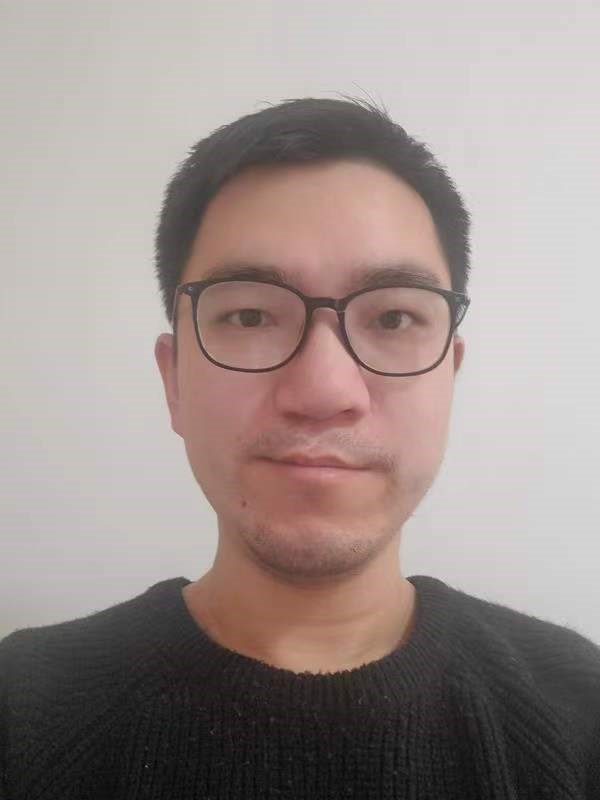}}]{Yujie He}
received the B.S. degree in Software Engineering from Xidian University,  Xi’an, China, in 2009,and the Ph.D. degree in Intelligent Information Processing from Xi’an Research Inst. of Hi-Tech, Xi’an, China, in 2016. 

His research interests include Infrared Small Target Detection, sparse signal representation, and infrared target tracking.
\end{IEEEbiography}

\end{document}